\documentclass[final,5p,times,onecolumn]{elsarticle}
\usepackage{amssymb}
\usepackage{rotating}
\usepackage{amsmath}
\usepackage{multirow}

\usepackage{tikz}
\usepackage{siunitx}
\usepackage{textcomp}
\usepackage{color,soul}

\usepackage[export]{adjustbox}
\usepackage[first=0,last=9]{lcg}
\usepackage{color, colortbl}

\usepackage{algpseudocode}
\usepackage[ruled,lined]{algorithm2e}
\usepackage{amsmath}
\usepackage{amsfonts}
\usepackage{amssymb}

\newcommand{\hlx}{\textcolor{black}}

\journal{Elsevier}

\begin{document}

\begin{frontmatter}
\title{Uncertainty Aware Neural Network from Similarity and Sensitivity}

\author{H M Dipu Kabir$^a$, Subrota Kumar Mondal$^b$,  Sadia Khanam$^c$, Abbas Khosravi$^d$,\\ Shafin Rahman$^e$, Mohammad Reza Chalak Qazani$^d$,  Roohallah Alizadehsani$^d$\\ Houshyar Asadi$^d$, Shady Mohamed$^d$, Saeid Nahavandi$^{d,f}$, U Rajendra Acharya$^{g,h,i}$\\
$^a$ Independent researcher.
$^b$FIE,  Macau University of Science and Technology, Macao.
$^c$Dhaka Dental College,\\ Dhaka, Bangladesh.
$^d$ IISRI, Deakin University, Australia.
$^e$North South University, Dhaka, Bangladesh.\\
$^f$Harvard University, Allston, MA 02134 USA.
$^g$ECE, Ngee Ann Polytechnic, Singapore.\\
$^h$BME, SST, SUSS University, Singapore.
$^i$BIME Asia University, Taiwan.\\
hmdkabir@connect.ust.hk
}

\begin{abstract}	
Researchers have proposed several approaches for neural network (NN) based uncertainty quantification (UQ). However, most of the approaches are developed considering strong assumptions. Uncertainty quantification algorithms often perform poorly in an input domain and the reason for poor performance remains unknown. Therefore, we present a neural network training method that considers similar samples with sensitivity awareness in this \hlx{paper. In} the proposed NN training method for UQ, first, we train a shallow NN for the point prediction. Then, we compute the absolute differences between prediction and targets and train another NN for predicting those absolute differences or absolute errors. Domains with high average absolute errors represent a high uncertainty. In the next step, we select each sample in the training set one by one and compute both prediction and error sensitivities. Then we select similar samples with sensitivity consideration and save indexes of similar samples. The ranges of an input parameter become narrower when the output is highly sensitive to that parameter. After that, we construct initial uncertainty bounds (UB) by considering the distribution of \hlx{sensitivity aware} similar samples. Prediction intervals (PIs) from initial uncertainty bounds are larger and cover more samples than required. Therefore, we train bound correction NN. As following all the steps for finding UB for each sample requires a lot of computation and memory access, we train a UB computation NN. The UB computation NN takes an input sample and provides an uncertainty bound. \hlx{The UB computation NN is the final product of the proposed approach.} Scripts of the proposed method are available in the following GitHub repository: https://github.com/dipuk0506/UQ

\end{abstract}

\end{frontmatter}

% \fi

\section{Introduction}
Whenever any prediction system \hlx{receives an unexpectedly high prediction error}, people try to investigate the reason for that failure. NNs consist of weights and biases. Values of those parameters are determined during the training. Therefore, the training procedure and the sample distribution play a vital role in NNs heteroscedastic performance \cite{theisen2021evaluating, fan2021sparse}. The knowledge of sample distribution also helps \hlx{individuals in the investigation}. Moreover, \hlx{NNs training needs to be} robust to achieve competitive performance on rare and critical samples, and a high overall performance \cite{koochali2021if, lakshminarayanan2017simple}.   

Traditional regressive \hlx{models can find the regression mean of a quantity} for any input combination within the range \cite{morala2021towards, mir2021neural}. They also provide an overall statistical error, such as the mean-square-error (MSE), the root-mean-square-error (RMSE), etc \cite{kavousi2020evolutionary}. Traditional point prediction models, with an overall statistical error, cannot represent the level of heteroscedastic uncertainty. The level of uncertainty can be high in one input domain and low in another input domain \cite{dong2021optimal, zare2022accurate}. Prediction intervals with a coverage probability can indicate the level of heteroscedastic uncertainty. \hlx{Regions with narrower intervals have lower uncertainty. Regions with wider interval have lower uncertainty} \cite{mobtahej2021effective}. \hlx{Uncertainty quantification is becoming popular in various fields \cite{ikidid2021multi, gomez2016uncertainty}. Neural networks are also getting popularity due to their optimal performances \cite{harish2020automated}.} There exist several popular approaches for constructing prediction intervals. Bayesian neural networks are getting popular due to their applications in deep learning. However, Bayesian regressive neural networks struggle to maintain a coverage probability, close to the expected coverage probability. The cost-function-based prediction intervals can provide a narrow interval and the coverage probability becomes very close to the expected coverage probability \cite{elder2021learning}. However, there are different cost functions proposed by different researchers, and there exist debates on their acceptability. A group of researchers prescribed reducing failure distance and some other researchers prescribed bringing the target near the mid-interval. There is also debate on penalizing high coverage probability. Therefore, we propose a method that trains NNs based on similar samples instead of a cost function. 

\begin{figure*} 
  \centering
  \includegraphics[clip, trim=1.5cm 6.2cm 6.6cm 2.2cm, width=6in,angle=0]{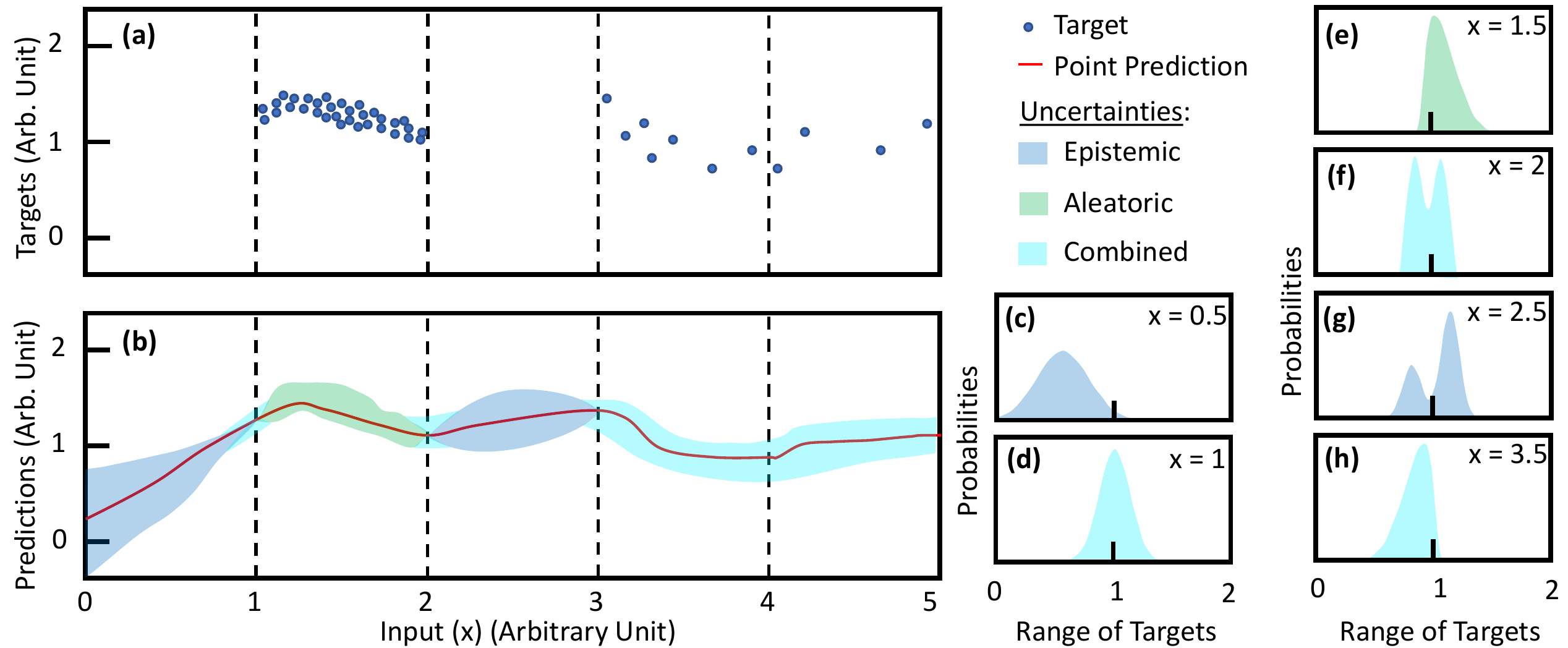}
  \caption{Rough sketches for the visualization of the distributions of uncertainties. Sub-sketch (a) presents the distribution of targets. Sub-sketch (b) presents aleatoric, epistemic, and combined uncertainties due to insufficient data as intervals. Sub-sketches (c) to (h) present probable probability distributions of uncertainties for several different values of inputs. }
  \label{UQ_sk}
\end{figure*}

\begin{figure} 
  \centering
  \includegraphics[clip, trim=0cm 0.0cm 0cm 0.0cm, width=3.5in,angle=0]{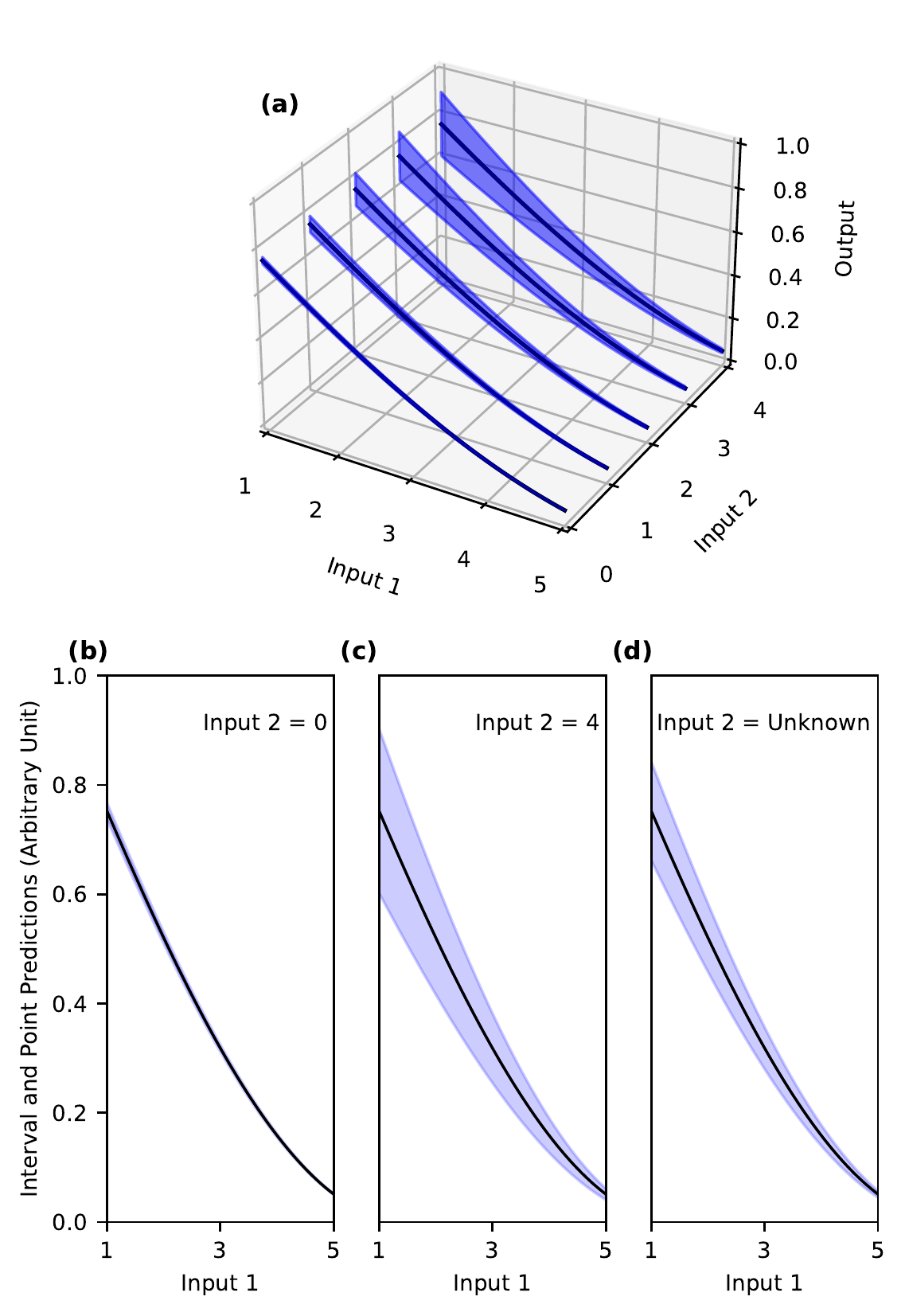}
  \caption{A parameter may not affect the prediction but may cause uncertainties. This figure shows rough sketches to visualize the influences of such parameters in uncertainty quantification. In subplot (a), \emph{Input 1} is responsible for both prediction and uncertainty. \emph{Input 2} is responsible for uncertainties. Subplots (b) and (c) present uncertainties for two different values of \emph{Input 2}. Subplot (d) presents the level of quantified uncertainty when \emph{Input 2} is unknown. }
  \label{Input_uncertainty}
\end{figure}

Although neural networks have brought promising results in numerous fields, they are prone to overfit in highly uncertain datasets \cite{goel2022directed, zilly2021plasticity}. Predictions from neural networks mainly face two types of uncertainties: (i) aleatoric uncertainty and (ii) epistemic uncertainty. Epistemic uncertainty is known as the modeling error \cite{malinin2018predictive}. NNs of optimal size can result in a much lower epistemic uncertainty. However, the presence of a high aleatoric uncertainty can potentially bring a high epistemic uncertainty as the NN may fail to evaluate the exact regression mean. Moreover, optimal size NN depends on both application and data. Shallow NNs with a few neurons can potentially exhibit the underfit issue due to insufficient parameters. Deeper NNs can potentially exhibit overfitting errors \cite{ma2022hw}. Trial and error is the only way to achieve the optimal size. Both the aleatoric and epistemic uncertainty may stay combinedly in many situations. The third type of uncertainty revels, while someone tries to present the uncertainty. That is representational uncertainty.

Fig. \ref{UQ_sk} presents rough sketches presenting aleatoric and epistemic uncertainties and their combined uncertainties. Fig. \ref{UQ_sk}(a) presents the distribution of targets. Fig. \ref{UQ_sk}(b) presents a rough interval prediction of the expected value. When the sample density is high with aleatoric variances, the NN mostly exhibits aleatoric uncertainties. In regions lacking samples, the prediction system exhibits epistemic uncertainty due to the lack of training examples. Regions with insufficient samples can potentially exhibit combined aleatoric and epistemic uncertainty. In Fig. \ref{UQ_sk} (b), rough diagrams are drawn for visualization. Combined uncertainty can also occur at the edge of two regions. There is an edge at x=1, joining a no-sample region and a high-density region. That edge has an uncertainty partially caused by the aleatoric variation in the high-density region and partially caused by the lack of example samples in the no-sample region.
As the width of the intervals represents the level of uncertainty, it is possible to understand the level of heteroscedastic uncertainties and a highly probable region of outcome from prediction intervals. However, representational uncertainty exists while presenting uncertainties as intervals. Intervals do not represent multimodality and skewness of uncertainty distributions. 

Fig. \ref{UQ_sk}(c)-(h) are presenting probability densities at different locations. Probability densities can provide a more detailed visualization of uncertainty. In Fig. \ref{UQ_sk}(c)-(d), probability distributions are purely Gaussian. In Fig. \ref{UQ_sk}(e) and Fig. \ref{UQ_sk}(h), probability distributions are skewed Gaussian. In Fig. \ref{UQ_sk}(e)-(g), probability distributions are multimodal. An interval contains three numbers, the upper bound, the lower bound, and the coverage probability. The prediction interval-based representation of uncertainties cannot express the skewness or multimodality, resulting in representation uncertainties. Moreover, real-world quantities are dependent on numerous factors \cite{ito2022backup, siljak2021artificial}. Drawing probability distribution over one variable is possible while i) other variables contain a certain set of values, or ii) other variables are unknown. Therefore, avoiding representational uncertainty is almost impossible in current representation approaches.

In a real-world scenario, many unknown parameters may cause uncertainty in the value of a quantity. The temperature in a location on a sunny day may face uncertainty depending on the presence of a cloud at that location. Fig. \ref{Input_uncertainty} presents a system containing two inputs. The first input is responsible for both prediction and uncertainty and the second input is partially responsible for uncertainty. A 3D plot in Fig. \ref{Input_uncertainty}(a) shows that the point prediction increases with the decrease of the first input. The magnitude of uncertainty varies based on the second input. The user receives more specific uncertainty information when the second input is known. Fig. \ref{Input_uncertainty}(b) presents a situation when the second input is zero. The uncertainty becomes very low in such a situation. Fig. \ref{Input_uncertainty}(c) presents a situation when the second input has its highest value. The uncertainty is much higher than Fig. \ref{Input_uncertainty}(b). Fig. \ref{Input_uncertainty}(d) presents uncertainty over the first input when the second input is unknown. That input may have any value between zero to four. Based on the value of the second input, there exist different levels of uncertainties. It is also possible to quantify uncertainty in absence of the second input by averaging the effect of different values of the second input. However, knowing a parameter related to uncertainty makes us more certain about the predicted value and makes intervals statistically narrower with expected coverage probability. When the first input is zero, modeling uncertainty without the information of the first input makes the interval wider than expected. When the first input is four, modeling uncertainty without the information of the first input results in intervals of low statistical coverage probability. 
\hlx{Cost function-based NN training methods automatically consider similarity, sensitivity, and error sensitivity when they are trained through a reward-based system. However, they have the following limitations: 1) there is no widely accepted cost function, and 2) we cannot observe similar samples. Similarity-based NN training methods do not face the debate of cost function. However, they do not consider error sensitivity and that results in an inferior performance. We are considering the error sensitivity for the first time which is not considered in similarity-based prediction systems.}

\hlx{This manuscript is the upgraded version of our previous conference paper \cite{kabir2020uncertainty}. We have improved the paper with the consideration of error sensitivity. Consideration of the error sensitivity has improved the performance. Besides performance improvement, we have shared many resources for future researchers. The scripts of the initial paper were developed on the Matlab software. We develop Python codes for the proposed method. As Python is more popular in data science, we expect that more researchers will use our code. We have also shared scripts of several other approaches. That may also help future researchers.}   

The rest of the paper is organized as follows. Section \ref{sec2} presents a short overview of uncertainty-aware machine learning.
Section \ref{sec3} presents the framework of the proposed method.
Section \ref{secdata} presents investigated datasets.
Section \ref{secresult} presents results with relevant texts, figures, and tables. Section \ref{secconc} is the concluding section.

\if 0
\section{Types of Uncertainties}
\subsection{Aleatoric}
\subsection{Epistemic}
\subsubsection{Model Uncertainties}
\subsubsection{Representational Uncertainties}
\fi

\section{Research on Uncertainty Aware Machine Learning} \label{sec2}
Researchers have developed numerous uncertainty-aware machine learning algorithms in recent decades \cite{abdar2021review, siljak2021artificial, ito2022backup}. There are two overlapping research terms in machine learning and uncertainty quantification: i) Uncertainty in ML \cite{gal2016dropout} and ii) ML for UQ \cite{kabir2018neural}. The first term represents the uncertainty in ML models. As deep learning models contain a large number of parameters and often receive a large input, such as high-resolution images and they are prone to overfit \cite{ajlan2022text, kamal2021alzheimer}. Both overfitting and underfitting cause epistemic uncertainty. The second term is popularly known as the NN-based uncertainty quantification. Many quantities in nature have partially random and partially deterministic portions. NN is applied to quantify uncertainties in those quantities. 
%Common terms in the field of deep uncertainty quantifications are as follows:
%\subsection{Common Terms in Uncertainty Quantification}

\if 0
\textbf{Bayesian inference:}
\textbf{Variational Inference:}
\textbf{Dropout:} 
\textbf{Ensemble:}
\fi

\subsection{Bayesian Techniques}
Bayesian techniques are popular approaches in Deep Uncertainty quantification. In a Bayesian interference, the user holds a prior belief about the previous state. The probability of a hypothesis is determined considering prior beliefs and the available observations. Bayes's theorem is applied to update the probability distribution for upcoming information. In Bayesian statistics, when we observe an event $x_E$, the probability of happening a hypothesis $y_H$ simultaneously is as follows:
\begin{equation}\label{Bayes}
P(y_H|x_E) = \frac{P(x_E|y_H)}{P(x_E)} P(y_H)
\end{equation}
where, $P(.)$ is the probability function. 
There exist several popular variants of Bayesian Learning. Such as Variational Inference \cite{campbell2019sparse}, Bayesian Active Learning \cite{zhou2021survey, olivier2021bayesian}, Dropout as Bayesian Approximation \cite{serpell2019probabilistic, posch2020correlated}, Laplacian Approximations \cite{jospin2022hands, upadhyay2021robustness}, etc.

%\subsubsection{Variational Inference}

%\subsubsection{Dropouts} 

%\subsubsection{Bayesian Active Learning}

%\subsubsection{Laplacian Approximations}

%\subsection{Ensemble Techniques}

%\subsubsection{Deep Ensemble}

%\subsubsection{Bayesian Deep Ensemble}

\begin{figure*} 
  \centering
  \includegraphics[clip, trim=1.2cm 4.8cm 5.7cm 4.8cm, width=4.5in,angle=0]{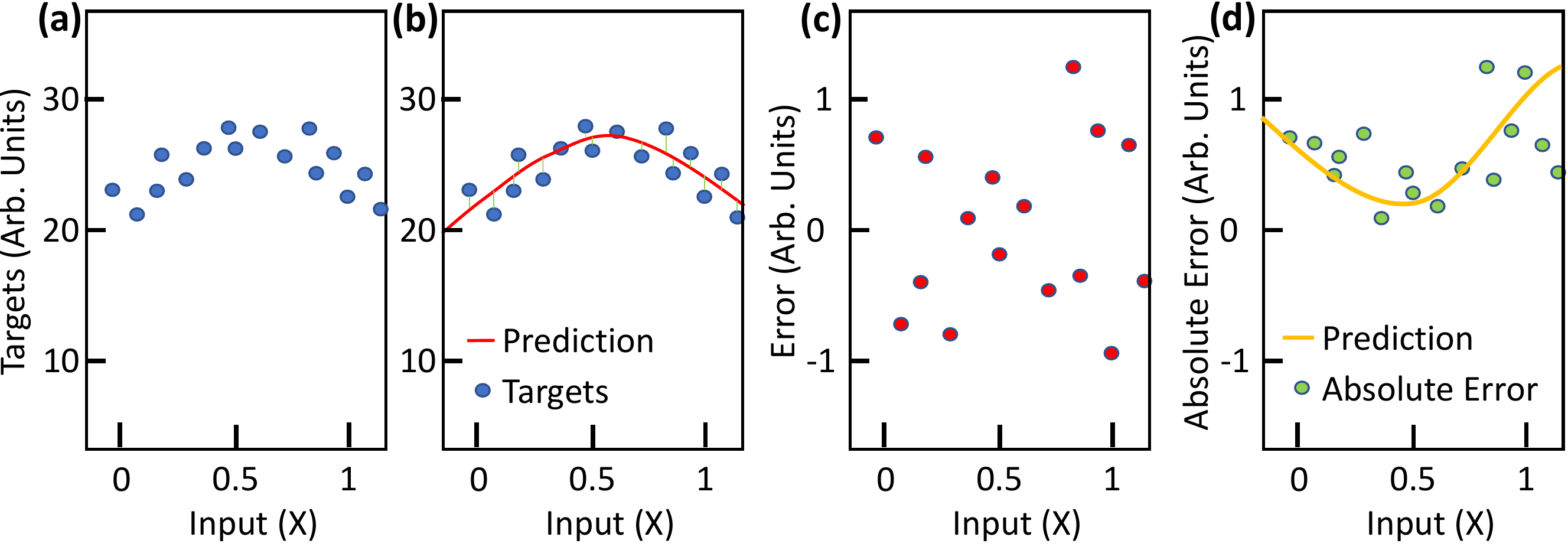}
  \caption{The reason for the development of error prediction NN on absolute error. An input parameter can be responsible for prediction or uncertainty, or both. Subplot (a) visualizes a target distribution over a range of one input.  (b) visualizes targets with predictions. (c) presents prediction errors with rough values for visualization. Subplot (e) presents the mean absolute error.  Uncertainty is high near 0 and 1. Uncertainty is low near 0.5. Prediction of error cannot represent the level of uncertainty where prediction of absolute error can predict that. }
  \label{UQ_AE}
\end{figure*}

\subsection{Cost Function-based Direct Interval}
Cost function-based direct interval construction neural network training is popularly used for uncertainty quantification in natural quantities. Well known examples are uncertainty quantification of wind power generations, electricity demand, electricity price, etc. The lower-upper bound estimation (LUBE) method is the very first and a highly cited cost function-based uncertainty quantification method \cite{khosravi2010lower}. The LUBE method constructs prediction-interval-based uncertainty quantification system. The LUBE method considers a combination of prediction interval coverage probability (PICP) and prediction interval normalized average width (PINAW). 

The PI normalized average width can be obtained as:
\begin{equation}\label{Eq:PINAW}
PINAW =   \frac{1}{R \times n} \sum_{j=1}^n (\overline{y}_j-\underline{y}_j),
\end{equation}
where, $\overline{y}_j$ is the upper bound for $j^{th}$ sample, $\underline{y}_j$ is the lower bound for $j^{th}$ sample, $R$ is the range of targets, $n$ is the number of samples. $PINAW$ provides a scalar value of the average width.
PICP is described as the following equation:
\begin{equation}
\label{eq:PICP}
PICP = \frac{1}{n} \sum_{j=1}^n c_j
\end{equation}
where,
\begin{equation*}
c_j = 	\begin{cases}
		1, t_j \in [\underline{y}_j, \overline{y}_j] \\
		0, t_j \not \in [\underline{y}_j, \overline{y}_j],
		\end{cases}
\end{equation*}
where, $t_j$ is the target for $j^{th}$ sample.
They developed the coverage width criterion CWC by combining PICP and PINAW as follows: 
\begin{equation}\label{CWC2}
CWC = PINAW \{ 1 + \gamma (PICP) e ^{\eta (1-\alpha - PICP)}\},
\end{equation}
where, $\gamma (PICP)$  is a penalty for low coverage ($PICP < 1-\alpha$). $\gamma (PICP)$ is represented by the following equation.
\begin{equation*}
\gamma (PICP) = 	\begin{cases}
		1, PICP < 1-\alpha \\
		0, PICP \geq 1-\alpha,
		\end{cases}
\end{equation*}
where, $\eta =50$ is a hyperparameter. $\alpha$ is the target non-coverage probability \cite{khosravi2011comprehensive}. After the proposal of the LUBE method, several researchers have proposed improved cost functions \cite{kabir2021optimal, marin2016prediction, zhang2015advanced} and more efficient training methods. Cost-function-based uncertainty quantification NN training methods have recently received a lot of attention in recent years due to their eye-catching performances.  However, their performances vary depending on considerations of parameters. A group of researchers concentrates on reducing the average PINAW, while another group aims to optimize PINAW and PICP in critical samples. Different cost functions are developed to achieve good coverage in critical situations. There exist confusion about the relative effectiveness of different cost functions.

\subsection{Effectiveness of NNs in Prediction and UQs}
Neural networks are the most extensively used predictors \hlx{in recent decades}. According to our literature search, researchers have been modeling engineering, scientific and economic quantities through mathematical equations for centuries. Developing a mathematical model requires observation of input-output relationships for different combinations of other inputs, mathematical knowledge of expressing relationships, and selection of parameters. Therefore, mathematical model development of complex systems requires skilled manpower and a long time for development.  

Many relationships between the input and outputs can be confusing to humans. The human brain can search over a limited number of events. To find a statistical relationship humans or machines may require to search over a large number of data. Such as the complex relationship between the temperature and the presence can potentially vary depending on several other factors. It is difficult for a skilled human to derive a mathematical relationship between cloud and temperature. Neural networks can easily find those statistical relationships through reward-oriented training.

Another popular and time-saving technique for the development of any prediction system is the lookup table. The lookup table method does not require extensive work from specialized humans. The prediction system interpolates the probable value from nearby samples. 
\hlx{However, the lookup table or similar samples requires large memory. Data loading time from memories also significantly increases.}
The memory requirement drastically increases with the number of input parameters. Moreover, datasets often lack an adequate number of samples in rare and critical domains. Neural network model parameters require less memory than the dataset. Training of NNs does not need extensive human work and the trained NN contains a much smaller chunk of data. Therefore, NNs are efficient in terms of both run-time memory and less human involvement.

\subsection{Similarity-based Uncertainty Quantification}
Several \hlx{researchers have demonstrated similarity-based prediction systems \cite{burnstein1961similarity, hiza1970correlation, yao2019revisiting, khan2011natural}. Researchers} can also obtain the level of uncertainty from similar samples \cite{vandecar1990determination}. However, such approaches did not get enough attention due to a high execution time. Model-based approaches require a large time during the model training but the execution is fast. Recent advancement in neural networks (NNs) has \hlx{brought state-of-the-art performances} in numerous fields. However, NN often fails with high confidence in critical situations. \hlx{The NN is a black box predictor. It is difficult to find the reason for any poor performance.} A potential solution can be efficient data pre-processing. It is possible to find aleatoric uncertainty and sample density from similar samples. An individual can train NNs on prediction and uncertainty values obtained from the similarity-oriented method. It is also possible to evaluate the statistical performance of trained models to compare performances.

\section{The Proposed Framework} \label{sec3}
%\subsection{Steps for UQ in Regression}

The proposed uncertainty quantification framework is computationally extensive during the data preparation for the NN training. However, the execution is faster, as a single input combination is applied to the trained shallow NN to obtain outputs. Besides the uncertainty-bound NN, we also train a sample density NN to help users. Sample density near a sample indicates the experience of the UQ NN near that sample. A very low sample density may cause a high epistemic uncertainty. Therefore, we facilitate users by providing a sample density neural network. In this paper, we propose a method containing the following steps for uncertainty quantification. We often receive data in a different format than the format required to train neural networks. Such as the original dataset may contain a string of samples taken at a certain interval with start and end times. The dataset for proposed NN training requires inputs(X) and output(Y) matrix. When the dataset is not originally in a proper format, the user may need some data pre-processing before applying the proposed steps. 

\begin{figure*} 
  \centering
  \includegraphics[clip, trim=0cm 0.0cm 0cm 0.0cm, width=5.7in,angle=0]{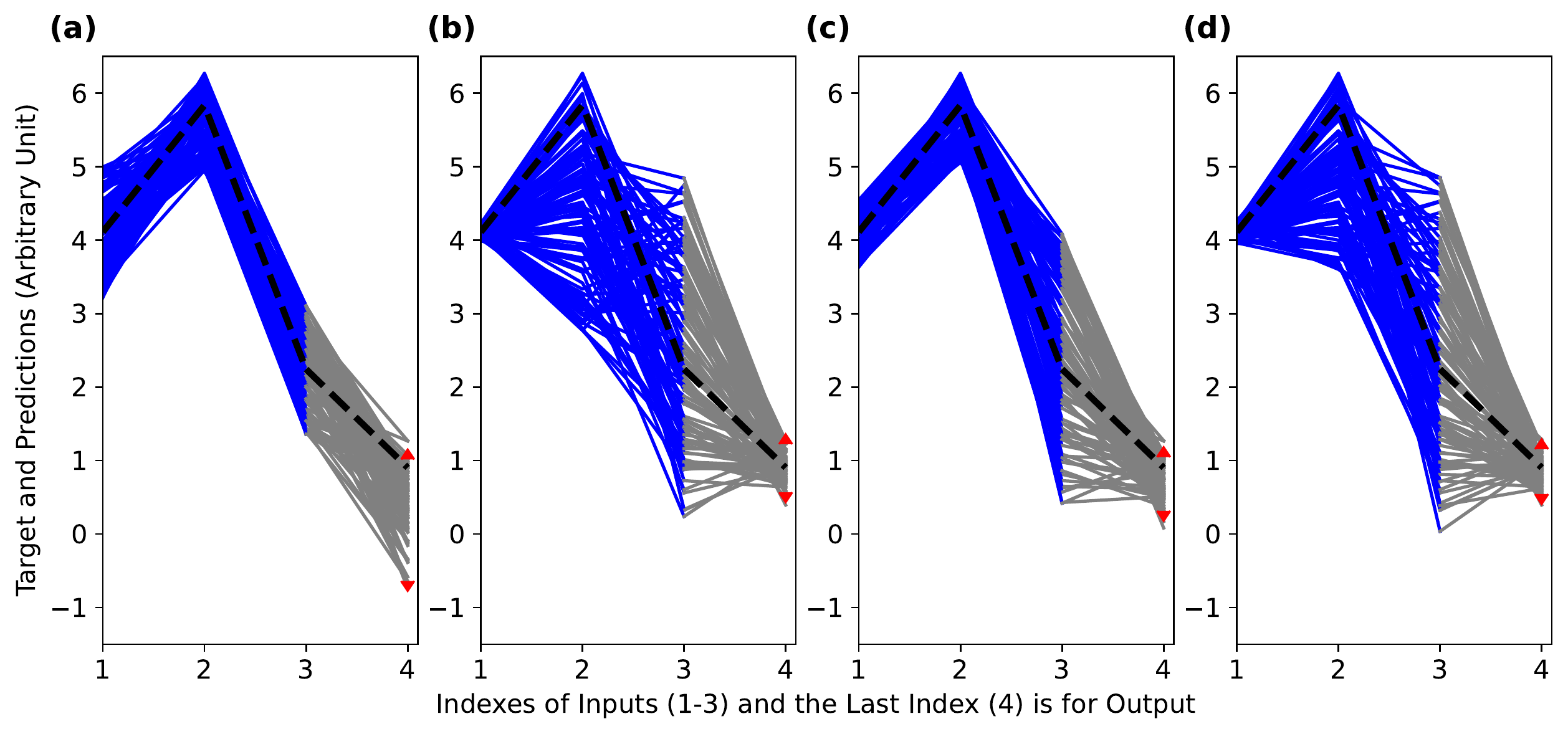}
  \caption{Similar samples with the example sample and derived rough prediction interval. Similar samples and prediction intervals are plotted while (a) only similarity, (b) similarity and prediction sensitivity, (c)-(d) similarity, prediction sensitivity and error-sensitivity. In (c), prediction sensitivity and error-sensitivity have equal weights. In (d), we adjust relative weights of prediction sensitivity and error-sensitivity based on relative variances. }
  \label{Similarity_plot}
\end{figure*}

\subsection{Find Similar Samples:}
Finding similar events is the most computationally extensive and critical part of the proposed system. The proposed method considers both error and prediction sensitivities while choosing similar samples. The following steps are followed to find similar samples from the dataset.

\subsubsection{Train a Shallow NN for Prediction:} We train an initial point prediction neural network and apply that neural network for obtaining both the prediction sensitivity and the error sensitivity.

\subsubsection{Compute Absolute Errors and Save:} The initially trained NN may have both aleatoric and epistemic uncertainty. However, that point prediction neural network cannot indicate the level of uncertainty. That neural network is applied to the entire training dataset to obtain the absolute error. Through that process, we obtain a new dataset containing input combinations and the absolute errors for these combinations.

Fig. \ref{UQ_AE} presents the reason for training NN for absolute errors instead of just errors.  Subplot (a) visualizes a target distribution over a range of one input.  Subplot (b) visualizes targets with predictions. Subplot (c) presents prediction errors with rough values for visualization. Subplot (e) presents the mean absolute error. 

An input parameter can be responsible for prediction or uncertainty, or both. Theoretically, the NN training brings the true-regression-mean. In a perfect situation, NN trained on the error signal should always predict a value very close to zero. Absolute errors are always positive. In less uncertain regions, absolute errors have small values. In a highly uncertain region, absolute errors have both small and high values and the true regression mean of absolute error value represents an indication of the level of uncertainty. In Fig. \ref{UQ_AE} uncertainty is high near 0 and 1. Uncertainty is low near 0.5. Therefore, prediction of error cannot represent the level of uncertainty where prediction of absolute error can predict that.

\subsubsection{Train a Shallow NN for Absolute Errors:}
The proposed method considers both the prediction sensitivity and the error sensitivity. The error contains both aleatoric and epistemic uncertainties. Moreover, one individual sample may not represent the actual level of uncertainty near that sample. A region of high uncertainty has a high statistical absolute error. Therefore, we are training a NN with absolute errors to achieve the regression mean of absolute errors. 

\subsubsection{Compute Sensitivity and Relative Weights:}We compute both prediction and error sensitivities from trained neural networks. Prediction sensitivity of $i^{th}$ input near $j^{th}$ sample can be computed as: 
\begin{equation}
\label{eq:sensitivity}
S_p(i)|_{X \rightarrow X_j} = \frac{\partial}{\partial x_i} NN_p(X)|_{X \rightarrow X_j}
\end{equation}
Consideration of sensitivity brings a superior performance compared to similarity alone \cite{kabir2020uncertainty}. The proposed work applies both prediction and error sensitivity to achieve superior performances.

\begin{algorithm}
	\caption{Obtaining Indexes of Similar Samples}
	\label{alg-sen}
	\small
	\text{\bf{Input} $\leftarrow$ Training dataset}\\
	\text{ }\\
	\text{\bf{Parameters and functions:}}\\
	\text{$N$ $\leftarrow$ Number of samples}\\
	\text{$n$ $\leftarrow$ Number of input parameters}\\
	\text{$Abs()$ $\leftarrow$ Absolute values}\\
	\text{$Var()$ $\leftarrow$ Variance}\\
	\text{$R_{Var}$ $\leftarrow$ Variance ratio}\\
	\text{$X$ $\leftarrow$ Inputs ($n$ $\times$ $N$ matrix)}\\
	\text{$T$ $\leftarrow$ Targets ($1$ $\times$ $N$ matrix)}\\
	\text{$i$ $\leftarrow$ Index of the corresponding sample}\\
	\text{$j$ $\leftarrow$ Index of the compared sample}\\
	\text{$R$ $\leftarrow$ Range of inputs ($1$ $\times$ $n$ matrix)}\\
	\text{$N_S$ $\leftarrow$ Number of close samples to select }\\
	\text{$NN_p$ $\leftarrow$ Point prediction NN }\\
	\text{$AE$ $\leftarrow$ Absolute errors ($1$ $\times$ $N$ matrix)}\\
	\text{$NN_e$ $\leftarrow$ Absolute error prediction NN }\\
	\text{$Dev_{Index}$ $\leftarrow$ [$Deviation$ $Index$] ($2$ $\times$ $N$ matrix)}\\
	\text{$Dev(i,j)$ $\leftarrow$ Deviation between two inputs ($1$ $\times$ $n$ matrix)}\\
	\text{$S_p$ $\leftarrow$ Prediction sensitivity ($1$ $\times$ $n$ matrix)}\\
	\text{$S_e$ $\leftarrow$ Error sensitivity ($1$ $\times$ $n$ matrix)}\\
	\text{$S_{en}$ $\leftarrow$ Combined sensitivity ($1$ $\times$ $n$ matrix)}\\
	\text{$I_{SS}$ $\leftarrow$ Indexes of selected similar samples ($N_c$ $\times$ $N$ matrix)}\\
	\text{$S_{Th}$ $\leftarrow$ Similarity threshold ($1$ $\times$ $n$ matrix)}\\
	
	\text{ }\\
%	\vspace{10}
	\text{\bf{Execution:}}\\
	\text{Train $NN_p$}\\
	\text{$AE$ $\leftarrow$ $Abs$\Big($NN_p$($X$) - $T$\Big)}\\
	\text{$R_{Var}$ $\leftarrow$ $\frac{Var(AE)}{Var(T)}$ }\\
	\text{Train $NN_e$}\\
	\For{$i$ $\leftarrow$ 1 to $N$}
    {
    Compute prediction sensitivity $S_p$ near sample $i$\\
    Compute error sensitivity $S_e$ near sample $i$\\
    $S_{en}$ $\leftarrow$ $S_e \times R_{Var} + S_p \times (1-R_{Var}) $\\
    \For{$j$ $\leftarrow$ 1 to $N$}
        {
        Compute $Dev(i,j)$\\
        $Dev_{Index}(j)$ $\leftarrow$ [$max(S_{en}. Dev(i,j)$./$R$) $\ \ \ $ $j$]
        }
        Sort ascending $Dev_{Index}$ based on deviation column \\
        $I_{SS}(i)$ $\leftarrow$ Indexes from First $N_c$ row from $Dev_{Index}$ matrix\\ 
        \Comment{Top $N_c$ matched indexes}\\
        $S_{Th}(i)$ $\leftarrow$ Maximum $S_{en}.Dev(i,j)$./$R$ among selected indexes \\
        
    }
    \text{\bf{Output} $\leftarrow$ $I_{SS}$, $S_{Th}$ }\\

\end{algorithm}

\subsubsection{Find Sensitivity Aware Similar Samples:}

Construction of prediction intervals from the sensitivity-aware selection of samples brings narrower intervals with expected coverage. Fig. \ref{Similarity_plot} presents the importance of sensitivity consideration with a proper ratio between error and prediction sensitivities. Fig. \ref{Similarity_plot}(a) presents a prediction interval construction technique that considers only similarity. Similar samples are selected based on value differences among similar parameters. As a result, selection intervals for all inputs are the same in Fig. \ref{Similarity_plot}(a).  Consideration of the similarity alone usually provides a wider interval. Some of the input parameters can be random to output. A random input parameter can make the interval wider as many very similar samples may have quite different values at the random parameter.  Also, a slightly less similar sample may have a closer value near the parameter which is random to output. Therefore, some closely related samples are discarded and some less closely related samples are considered. That causes output distribution more spread, causing a wider interval.

The algorithm in Fig. \ref{Similarity_plot}(b) considers both similarity and the prediction sensitivity. The algorithm with only prediction sensitivity consideration is available in \cite{kabir2020neural}. The first two inputs are responsible for the prediction and uncertainty respectively. The third input is random. The algorithm allows lower deviation in highly sensitive samples. Allowing very high deviation to an input parameter that is random to output keep similar samples under consideration. Therefore, interval in Fig. \ref{Similarity_plot}(b) is narrower than the interval in Fig. \ref{Similarity_plot}(a).

Fig. \ref{Similarity_plot}(c) presents similar samples and constructed prediction intervals while considering prediction and error-similarities equally. Fig. \ref{Similarity_plot}(d) presents similar samples and constructed prediction intervals while considering prediction and error-similarities based on their variation ratio, mentioned in Algorithm \ref{alg-sen}. Approaches with sensitivity consideration, shown in Fig. \ref{Similarity_plot}(b)-(d). Consideration of sensitivity bring narrow intervals while Fig. \ref{Similarity_plot}(d) becomes slightly narrower than Fig. \ref{Similarity_plot}(b) and Fig. \ref{Similarity_plot}(c). The probable reason for superior performance is a better trade-off between prediction and error sensitivity.

\hlx{The sensitivity of input varies based on the current value of parameters. For example, the effect of a slight time change on electricity demand depends on the current time and the current electricity demand value. The electricity demand changes rapidly during the evening. However, the electricity demand does not change rapidly during midnight. }

\subsection{Sample Density NN Training:}
While searching for similar samples according to Algorithm \ref{alg-sen}, we track the sensitivity-aware deviation of selected samples. The highest value deviation among selected samples is recorded and using that deviation, we obtain the sample density. The sample density provides a rough idea about the experience of NN near the sample. We train a NN for sample density prediction so that users do not require to search similar samples for finding sample densities. Users can compute a rough sample density near a sample by providing the sample input combination to the sample density neural network.

\subsection{Plotting Similar Events Near a Sample:}
Although the algorithm extracts similar samples for all samples in the dataset in the first subsection of this section. We have provided a separate script for observing similar samples for a specific sample in the GitHub repository for better visualization. The name of the script is \emph{A3- Plotting Similar Events.ipynb}. The user of the proposed prediction system can easily observe similar samples of a training sample with the help of that script.

\subsection{Bound Correction NN Training}
\begin{figure} 
  \centering
  \includegraphics[clip, trim=5.4cm 5.4cm 13cm 1.0cm, width=3.5in,angle=0]{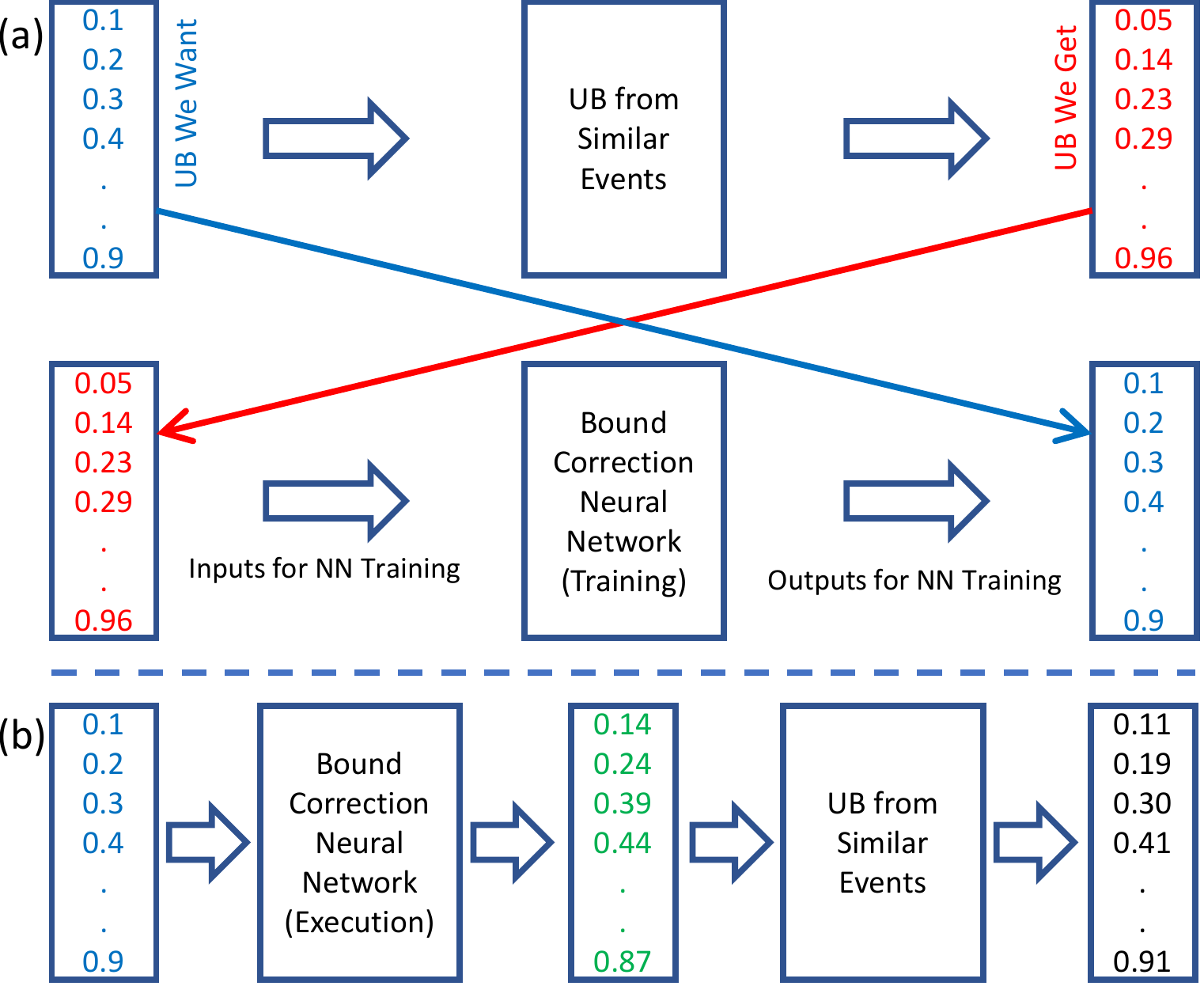}
  \caption{The bound correction NN training and execution. Although similar events have very close values of highly sensitive parameters, similar events are not the exactly same event. Therefore, prediction intervals from similar events are slightly wider than actual and covers more samples than expected. Subplot (a) visualize the process of bound correction NN training. UB we want is varied from 0.01 to 0.99 with 0.01 step. Subplot (b) shows how bound correction NN helps us with modified UB values.}
  \label{Bound}
\end{figure}

Intervals obtained from the distribution of similar samples are wider than required. As similar samples are not the same sample, outputs of similar samples are more spread. Fig. \ref{Bound}(a) presents the situation of distribution spreading and bound correction neural network training process. We select bound by sorting outputs of similar samples and by applying the initial UB value. We evaluate practical statistical bound-value for each UB value by investigating them on the entire training dataset. When we want to get an uncertainty lower bound of 0.10, we get an uncertainty lower bound of about 0.05. Lower bounds are lowered due to the extra spread. Upper bounds shifts upwards due to the same reason. If we want to get a lower bound of 0.05, we have to provide 0.10 to similarity count system. Therefore, we develop a bound correction NN which converts 0.05 to 0.10. Fig. \ref{Bound}(b) presents how we apply the bound correction NN to get optimal UB values.

\begin{figure*} 
  \centering
  \includegraphics[clip, trim=2.4cm 12.9cm 1.8cm 1.5cm, width=6.7in,angle=0]{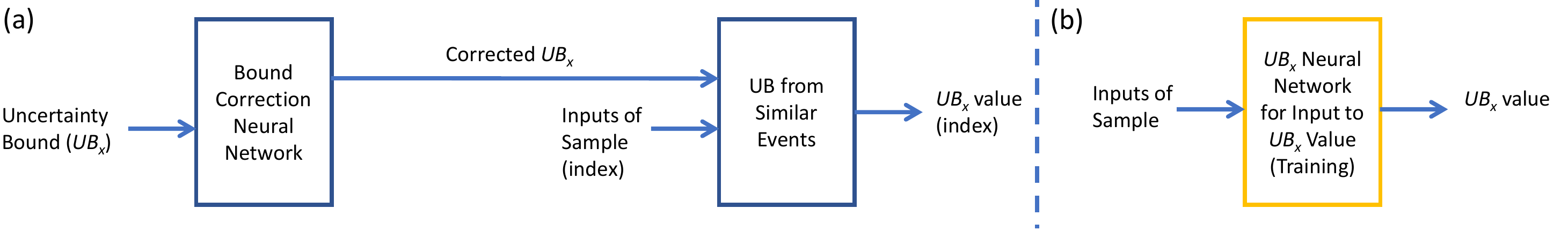}
  \caption{Reason for the direct NN for UB. (a) presents the proposed detailed method of computing uncertainty bound. The detailed method consists of bound correction, finding similar events from the entire test data, and finding the bound from distributions. The process is computationally extensive, and the system requires access to the entire training data for computing the UB. Therefore, we compute UB of certain cumulative probability ($UB_{CP}$) through the detailed method and use those input-output combinations for training a direct NN. (b) shows the structure of the direct NN of a certain cumulative probability $UB_{x}$.}
  \label{DirectNN}
\end{figure*}

\subsection{Direct NN Training for UQ}

Obtaining PI from similar samples requires the dataset. Computing predictions after searching the entire data is not feasible. Therefore, we train a bound prediction NN based on the bound-corrected uncertainty bounds.  Fig. \ref{DirectNN}(a) presents the reason for training a NN for direct uncertainty quantification. The traditional process requires finding similar samples, bound correction NN, and obtaining the bound from the distribution of samples. That process is computation extensive and memory hungry. Therefore, we follow these steps during the training and obtain uncertainty bound for the entire training dataset. We train NNs for training input combination and obtained uncertainty bounds as example outputs. The direct NN with input-output combination is available in Fig. \ref{DirectNN}(b). During the execution, a single input pattern and an uncertainty bound are provided as the input of NN. The user of the NN can also change the value of uncertainty bound ($x$) and train another NN. Through this approach, the user can obtain multiple uncertainty bounds and intervals.

Although it is possible to train a NN containing both the varying UB and default input combinations, we investigate NNs with a single UB ($UB_{CP}$). We observe high epistemic uncertainty while developing a single NN for the prediction of different UB values. Therefore, we train different NN for different UB values and that significantly lowers the epistemic uncertainty.  

\if 0
\subsection{Steps for Classification}

\begin{figure*}
\begin{center}
\includegraphics[clip, trim=1.2cm 12.2cm 8.0cm .3cm, width=14cm] {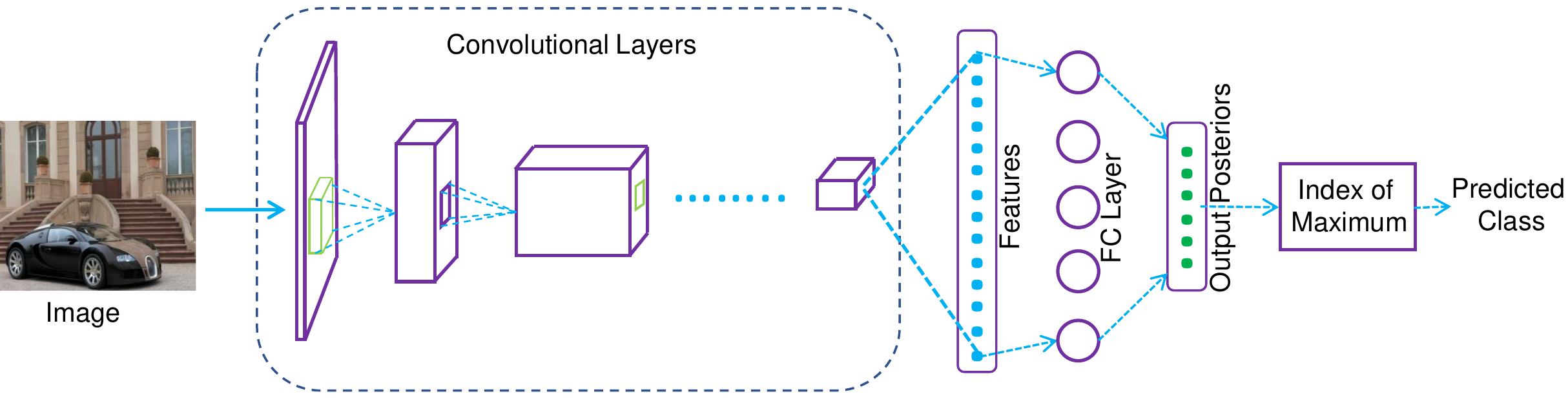}
\caption{\label{Posterior} A deep convolutional NN for classification problems. The convolutional layers of NN receive an image as input and generate features. A fully connected (FC) layer receives those features and generates output posteriors. Indexes of the maximum value of output posterior become the predicted class number. Output posteriors are also useful in computing running loss during the training. }
\end{center}
\end{figure*}
\fi

\hlx{Table \ref{CompMethod} presents comparative merits and demerits of different uncertainty quantification methods. Point prediction with the root-mean-square-error (RMSE) or any variance information provides prediction intervals of constant width and does not reflect heteroscedasticity.  Delta, Bayesian, mean-variance-estimation (MVE), and Bootstrap methods consider heteroscedasticity, however, fail to capture asymmetry due to the assumption of Gaussian distribution. Lower-upper-bound estimation (LUBE) method brings a smart prediction interval with the consideration of asymmetry and heteroscedasticity. The LUBE method does not follow any assumption on the distribution.  However, the LUBE method is a cost function-based NN training method. Therefore, the LUBE method cannot track similar samples. Common similarity-based methods often perform poorly, as they do not consider sensitivity. The proposed method considers both similarity and sensitivity to construct optimal uncertainty bounds.}

\begin{table}
\centering
\caption{Uncertainty estimation approaches and their characteristics. }
\label{CompMethod}
\begin{adjustbox}{width=3in,center}
\begin{tabular}{|c|c|c|c|c|c|}
\hline
Approach & \multicolumn{4}{|c|}{Merits }  & \\ \cline{2-5}
         & \rotatebox{90}{Heteroscedasticity }  & \rotatebox{90}{Asymmetry} & \rotatebox{90}{Observe Similar Samples } & \rotatebox{90}{Considers Sensitivity } & \rotatebox{90}{Assumption on Distribution}   \\ \hline 
 Point Prediction with RMSE & No & No & No & Yes & Yes \\ \hline 
 Delta Method & Yes & No & No & Yes & Yes \\ \hline
 Bayesian Method & Yes & No & No & Yes & Yes \\ \hline
 MVE Method & Yes & No & No & Yes & Yes \\ \hline
 Bootstrap Method & Yes & No & No & Yes & Yes \\ \hline
 
 LUBE Method & Yes & Yes & No & Yes & No \\ \hline
 Similarity-based Methods & Yes & Yes & Yes & No & No \\ \hline
 Proposed Method & Yes & Yes & Yes & Yes & No \\ \hline

\end{tabular}
\end{adjustbox}
\end{table}

\section{Datasets Investigated} \label{secdata}
\begin{figure*}
  \centering
  \includegraphics[clip, trim=2cm 0.09cm 2.5cm 0.6cm, width=6.3in,angle=0]{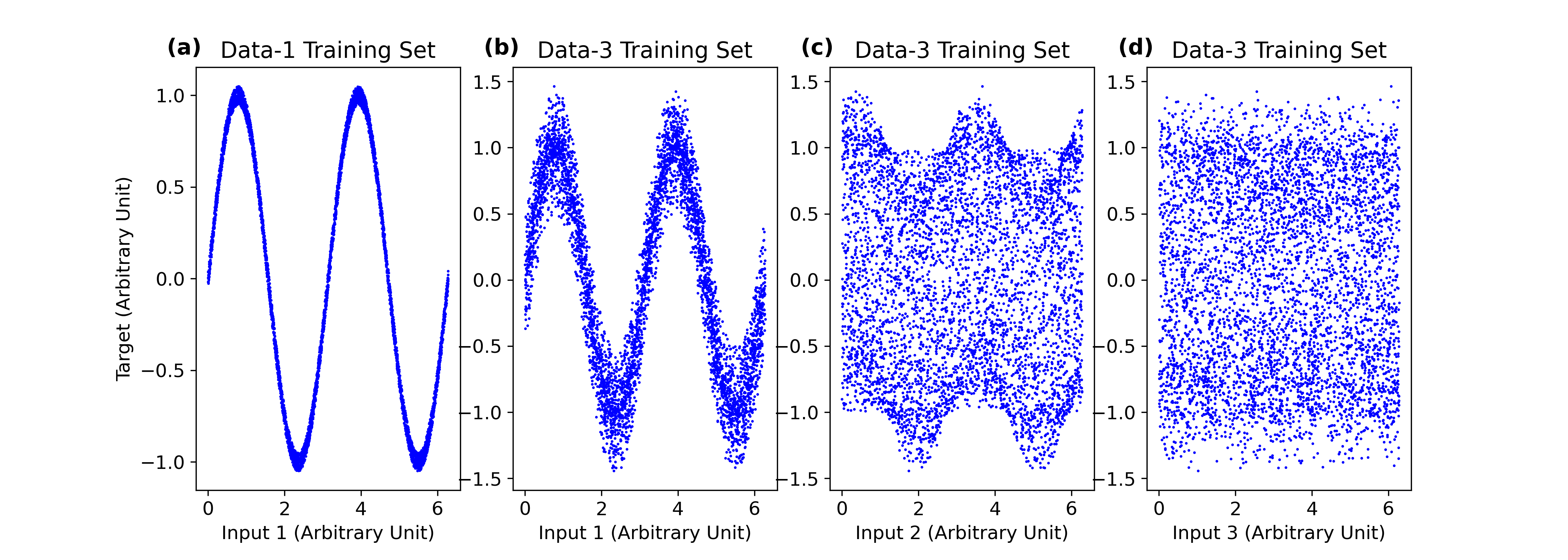}
  \caption{Relationship between target and inputs on Dataset-1, and Dataset-3\cite{kabir2023synthetic}. The first input highly influences the point prediction. The second input highly influences the level of uncertainty. The target is not related to the third input. }
  \label{Data3}
\end{figure*}

We investigate our proposed algorithm on both synthetic and real-world datasets. At first, we apply the proposed method to Dataset-1 and Dataset-3 of the synthetic data repository \cite{kabir2023synthetic}. Fig. \ref{Data3} visualizes input-output relationships in investigated synthetic datasets. Dataset-1 has one input and homoscedastic uncertainty over the input range.  Fig. \ref{Data3}(a) presents input-output relationship in Dataset-1. As there exists one input in Dataset-1, one figure can explain the input-output relationship. Fig. \ref{Data3}(b)-(d) present input-output relationships in Dataset-3. In this dataset, the first input determines prediction, the second input determines uncertainty, and the third input is random to output. A more detailed description of synthetic datasets is available in the paper \cite{kabir2023synthetic}. 

\begin{figure*}
  \centering
  \includegraphics[clip, trim=0cm 0.0cm 0cm 0.0cm, width=5.5in,angle=0]{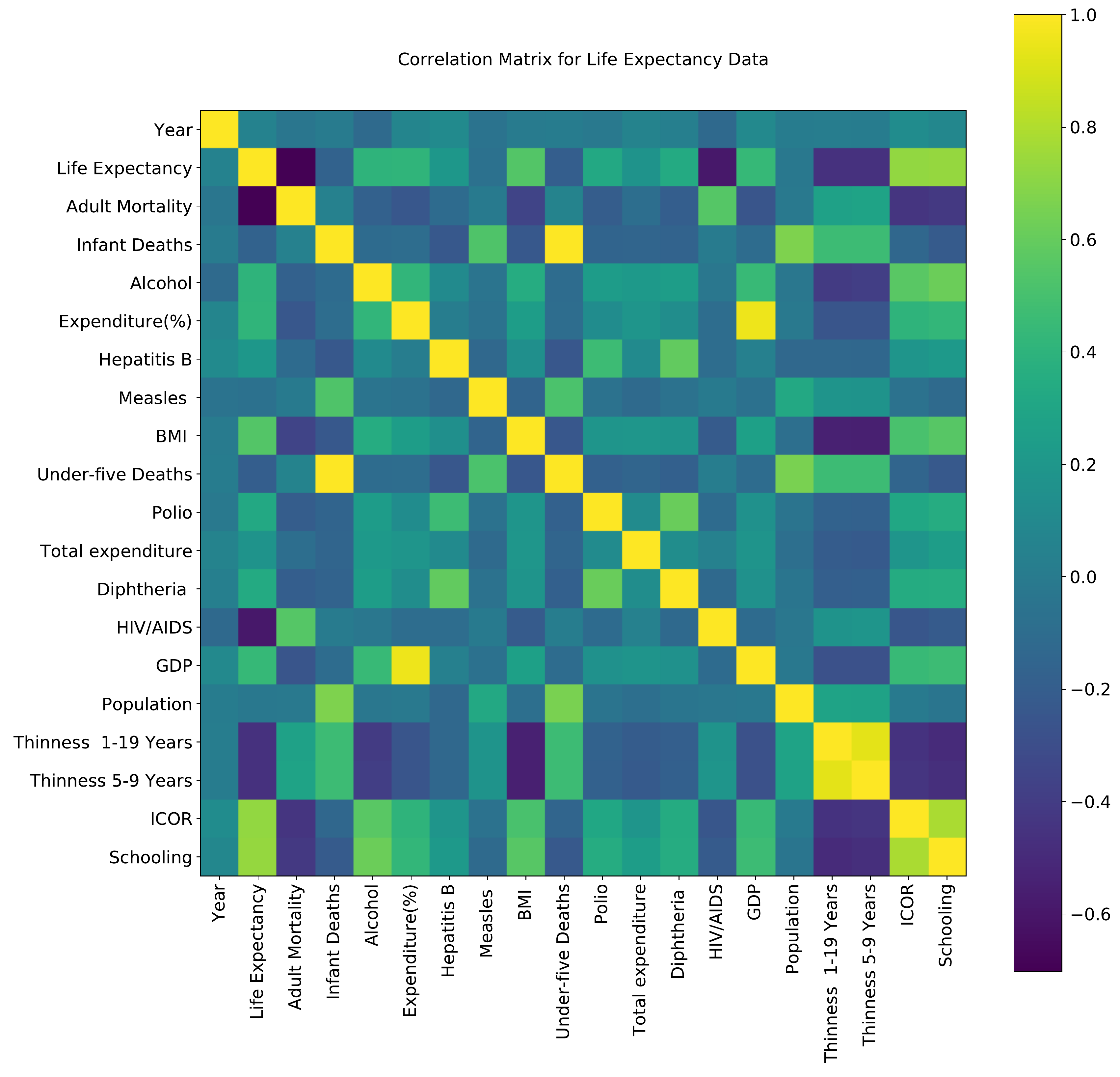}
  \caption{Correlation matrix on Global Health Observatory (GHO) data (2000-2015) \cite{LE_DATA}. The Dataset contains twenty numeric parameters and two non-numeric parameters. This figure shows a correlation matrix among numeric parameters. Non-numeric parameters are containing country names and development statuses. }
  \label{Cmat_LE}
\end{figure*}

Besides the initial investigation of the proposed method on synthetic datasets, we apply the method to Global Health Observatory (GHO) data (2000-2015) \cite{LE_DATA}. That dataset was collected from the World Health Organization (WHO) and United Nations (UN) websites and shared in the Kaggle platform \cite{LE_DATA}. That dataset contains a Comma-Separated Values (CSV) file. The CSV file contains a matrix. The matrix contains twenty-two columns. Two columns out of twenty-two columns contain texts: one contains country names and the other one contains the developing status of the country. The other twenty columns contain numeric values. A correlation matrix between numeric values is presented as Fig. \ref{Cmat_LE}.  The dataset contains 2938 entries. However, some entries lack one or more quantities. Discarding incomplete entries, there exist 1649 entries on the dataset. 
The correlation matrix provides a rough correlation between two parameters. Such as life expectancy has slightly increased over years during this period. Although most of the relationships are close to neutral, we can observe several high positive statistical correlations between parameters. The Gross Domestic Product (GDP) has a good statistical correlation with the percentage of expenditure. The number of infant deaths is also highly correlated to the number of deaths under-five years. The dataset provides the Income Composition of Resources (ICOR). ICOR is an index computed through education, life expectancy, per capita income, etc. ICOR has high positive statistical correlations with life expectancy, Body mass index (BMI), and schooling. This index is used to rank countries. Adult mortality has a statistically positive correlation to the number of AIDS patients. Several parameters are negatively correlated to each other. Life expectancy negative correlation with adult mortality, AIDS, and thinness at early ages. However, the correlation matrix is unable to grab periodic relationships. Fig. \ref{Data3}(a) shows a periodic relationship between input and output. NN can capture such periodic relationships where the correlation matrix may exhibit a near-zero correlation value.

We also investigate the performance of the proposed method on the electricity demand and wind power datasets. We download wind power generation and electricity demand samples from August 2012 to August 2022 from the UK-grid website \footnote{www.gridwatch.templar.co.uk/}. For both the Wind Power and Electricity Demand datasets, we consider four recent samples, time of day, and the day number of the year as inputs. We write a separate script for the data preparation. We also upload the data preparation script to the GitHub repository.

\hlx{Among our investigated datasets, only Dataset-1 has a constant standard deviation. Dataset-3 seems to have a constant standard deviation when they are observed in terms of the first input in Fig \ref{Data3}(b). However, Dataset-3 seems to have variable standard deviation or heteroscedastic uncertainty over the second input, shown in Fig \ref{Data3}(c). In Fig. \ref{Data3}(c), when the value of the second input stays within the range 1 to 3 inclusive, the upper range of output remains constant, and the lower range varies between -1 to about -1.5 based on the values of the second input. Besides synthetic datasets, we apply the algorithm to electricity demand, wind power, and life expectancy datasets. These three real-world datasets also have heteroscedastic uncertainties. In the electricity dataset, both the demand and the level of uncertainty are low during midnight, and they become high during the afternoon. Uncertainty in wind-power generation is also high when the wind speed is medium and varies over time. The wind-power generation becomes more certain when wind speed is either lower than the cut-in speed or higher than the rated speed. The life expectancy dataset also contains a complex relationship between input combination and uncertainty in targets. As those complex heteroscedastic relationships cannot be expressed in simple equations, we train NNs to predict them. }

\section{Results} \label{secresult}
We have applied different approaches to datasets. We applied Bayesian NNs from two different sources. Also, we apply the direct interval construction method with three different cost functions. We also apply the proposed approach with and without error sensitivity consideration.
We train NNs and investigate performances multiple times. As BNN models do not train a high performing prediction interval, we train BNNs for five times ($N_{Trial} =5$) and obtain statistical results. For cost-function-based direct interval methods and similarity based data preparation methods we train NNs for 100 times and obtain statistical results.

\begin{figure*}
  \centering
  \includegraphics[clip, trim=0.7cm 10.5cm 0.3cm 1.3cm, width=6.9in,angle=0]{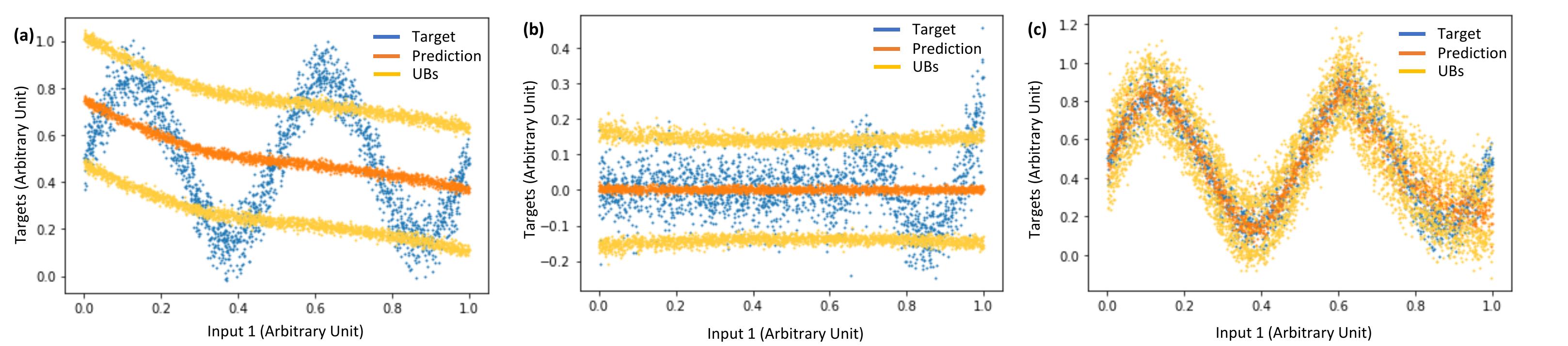}
  \caption{Point prediction and prediction intervals with Bayesian NN. (a) Pyro Bayesian NN. (b) Prediction and interval values while applying Pyro BNN to the error signal. (c) Prediction and interval values with targets after adjusting point prediction. }
  \label{Data3BNN}
\end{figure*}

\subsection{Bayesian Neural Networks}
We follow the scripts of Blundell et al. \cite{blundell2015weight, Pyro} and Lee et al. \cite{lee2022graddiv} as examples of Bayesian NN to quantify uncertainties. Scripts of Blundell et al. \cite{blundell2015weight, Pyro} is popularly known as BNN in Pyro. Scripts of Lee et al. \cite{lee2022graddiv} is popularly known as the TorchBNN module.

\subsubsection{Bayesian Neural Networks in Pyro}
The Pyro NN assumes a gaussian distribution of probability and provides regression mean and standard deviation. Using the mean value and standard deviation value, it is possible to draw intervals using the following relationship:
\begin{equation}
PI = [\mu -z \sigma, \mu + z \sigma]
\end{equation}
where, $\mu$ is the median and $\sigma$ is the standard deviation of the probability distribution. $z$ is a parameter which determines the statistical coverage probability of interval. $z$ becomes 1.15, 1.64, and 1.96 for 75\%, 90\%, and 95\% coverage probabilities respectively.
The investigated Pyro neural network contains two hidden layers. Each layer contains five hundred neurons. Pyro NNs are trained over twenty-five iterations. The customized Adam optimizer for pyro is applied for the optimization. The learning rate is set to 1e-3. According to our simulations, the Pyro UQ method overestimates the uncertainty. The intervals becomes much wider than required in the Pyro method. The first segment of Table \ref{TABComp1} presents the performance of the Pyro BNNs in investigated datasets. The first three rows of the table present the performance of Pyro intervals on  Dataset-1. We construct intervals of 95\%, 90\%, and 80\% overage probabilities. Constructed PIs are wider than required and covers more sample. As a result PICP values are much larger than required.

The default Pyro model cannot find the complex relationship between inputs and output with a reasonable statistical error. Fig. \ref{Data3BNN}(a) presents targets with Pyro mean and standard deviation curves. Neural network training methods often treat high-frequency changes as noise. Some training methods can capture 3-4 ups and downs over the range of one input but fails to be fitted for more than 5 ups and downs. However, human eyes can easily detect those patterns. Such as traditional NNs with proper training can capture trends of Dataset-7 \cite{kabir2023synthetic}, however, they fail to capture Dataset-8 of the synthetic dataset \cite{kabir2023synthetic}.

\subsubsection{Pyro Model with Shallow Point Prediction NN}
The Pyro model is unable to capture 3-4 ups and downs of the output for an input range. Therefore, the PINAW becomes several times larger than required and PICP becomes larger than required. Therefore, we apply a point prediction model to predict the regression mean of the dataset. Then we compute the error signal of the traditional error prediction model. Finally, we train a Pyro model to compute the uncertainty of the error signal. The prediction NN contains two hidden layers. Both layers contain one thousand neurons. Traditional NN training also applies the Adam optimizer with a learning rate of 0.05. Traditional NNs are trained over one thousand epochs. The training and network details of the pyro NN are the same as described in the previous subsection. Fig. \ref{Data3BNN}(b) presents the error signal. It also presents the regression mean and standard deviation interval of the error signal from the Pyro model. Fig. \ref{Data3BNN}(c) presents the targets with point prediction-adjusted Pyro outputs. Observing Fig. \ref{Data3BNN}(a) and Fig. \ref{Data3BNN}(c), it is obvious to us that Pyro with the help of a point prediction neural network provides superior performance compared to Pyro alone.

The second segment of Table \ref{TABComp1} presents the performance of predicted intervals on different datasets. PIs become sharper and yet covers the required percentage of samples. 

\subsubsection{Bayesian Neural Networks in TorchBNN Module}

\begin{figure*}
  \centering
  \includegraphics[clip, trim=0.7cm 3.6cm 0.3cm 1.3cm, width=6.9in,angle=00]{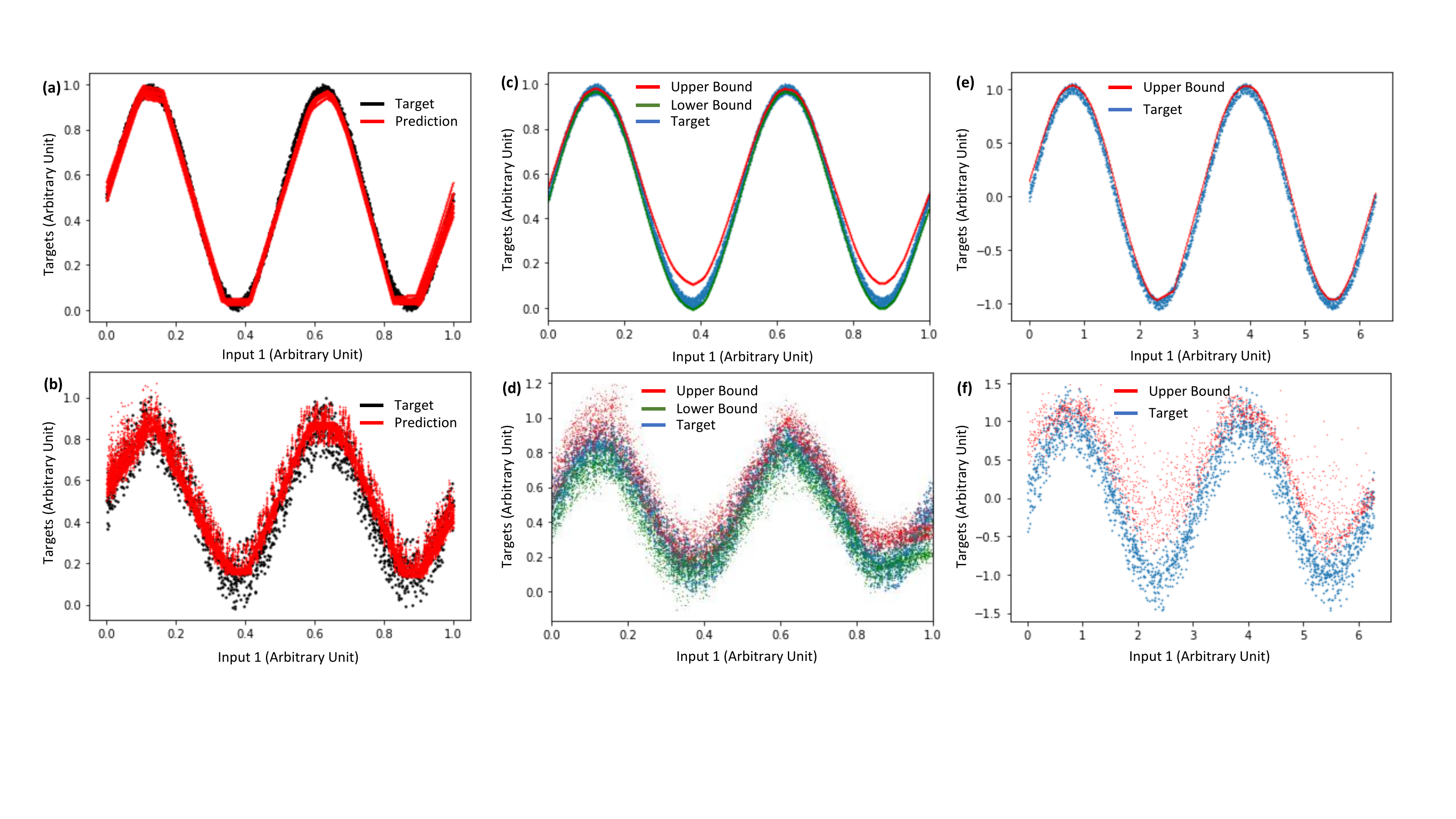}
  \caption{(a)-(b)Targets and predictions while applying the TorchBNN module on (a) Dataset-1 and (b) Dataset. The TorchBNN module provides a point prediction value instead of providing an interval prediction. However, as the parameters of the model are not like parameters of traditional neural networks, the prediction is slightly different from time to time, reflecting the uncertainty.
  (c)-(d) Targets with uncertainty bounds of 95\% coverage probability (c) on Dataset-1 (d) on Dataset-3 while applying the direct interval construction method with the optimal cost function.
  (e)-(f)Targets with uncertainty upper bounds of 90\% coverage probability while applying the proposed similarity, sensitivity, and error sensitivity-based method. The method constructs an uncertainty bound of a certain cumulative probability value. These figures show denormalized values. Therefore, the ranges in both axis are different from the normalized range (0 to 1). }
  \label{DataUB}
\end{figure*}

The TorchBNN module is developed by Lee et al. \cite{lee2022graddiv}. The TorchBNN module does not predict an interval or regression mean with standard deviation. Instead, it predicts a single value. That predicted value changes from time to time. The bayesian NN in the TorchBNN module is developed with two hidden layers. Neurons are declared with a prior $\mu$ of zero and a prior $\sigma$ of 0.1. Each layer contains five hundred neurons. We apply a cost containing both MSE loss and Keras variational autoencoder loss. The learning rate is set to 0.01. We train the TorchBNN BNNs over ten thousand iterations.  Fig. \ref{DataUB}(a)-(b) presents targets and predictions while applying the TorchBNN module on (a) Dataset-1 and (b) Dataset-3 \cite{kabir2023synthetic}. At first, the TorchBNN neural networks are trained on these datasets. Then, NNs are applied to the test data subsets. Targets and predictions are shown in Fig. \ref{DataUB}. 

\if 0
\begin{figure}
  \centering
  \includegraphics[clip, trim=20.6cm 3.5cm 2.5cm 1.4cm, width=2.8in,angle=0]{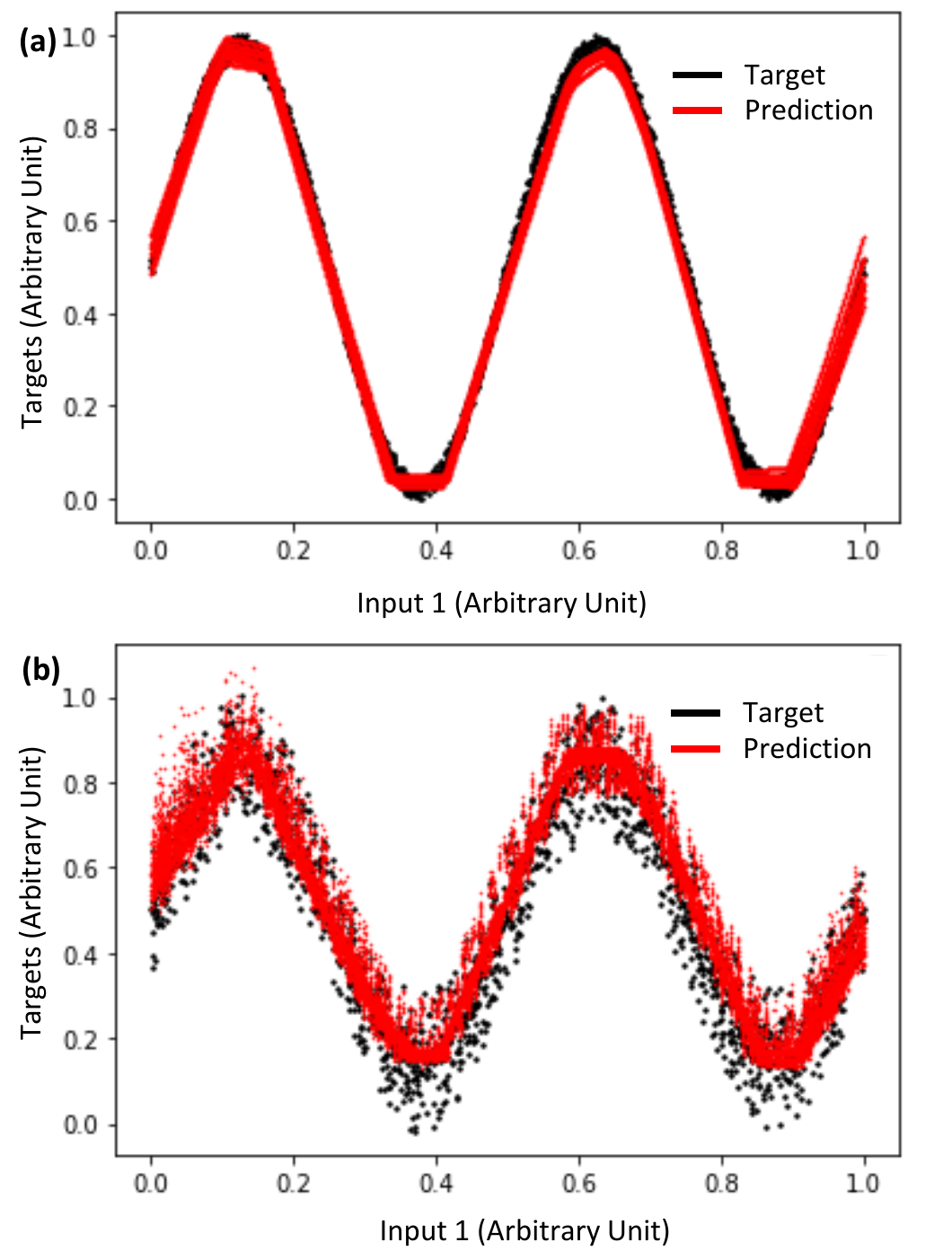}
  \caption{Targets and predictions while applying the TorchBNN module on (a) Dataset-1 and (b) Dataset. The TorchBNN module provides a point prediction value instead of providing an interval prediction. However, as the parameters of the model are not like parameters of traditional neural networks, the prediction is slightly different from time to time, reflecting the uncertainty.  }
  \label{Data3BNN2}
\end{figure}

\begin{figure}
  \centering
  \includegraphics[clip, trim=20.6cm 3.6cm 2.5cm 1.4cm, width=2.8in,angle=0]{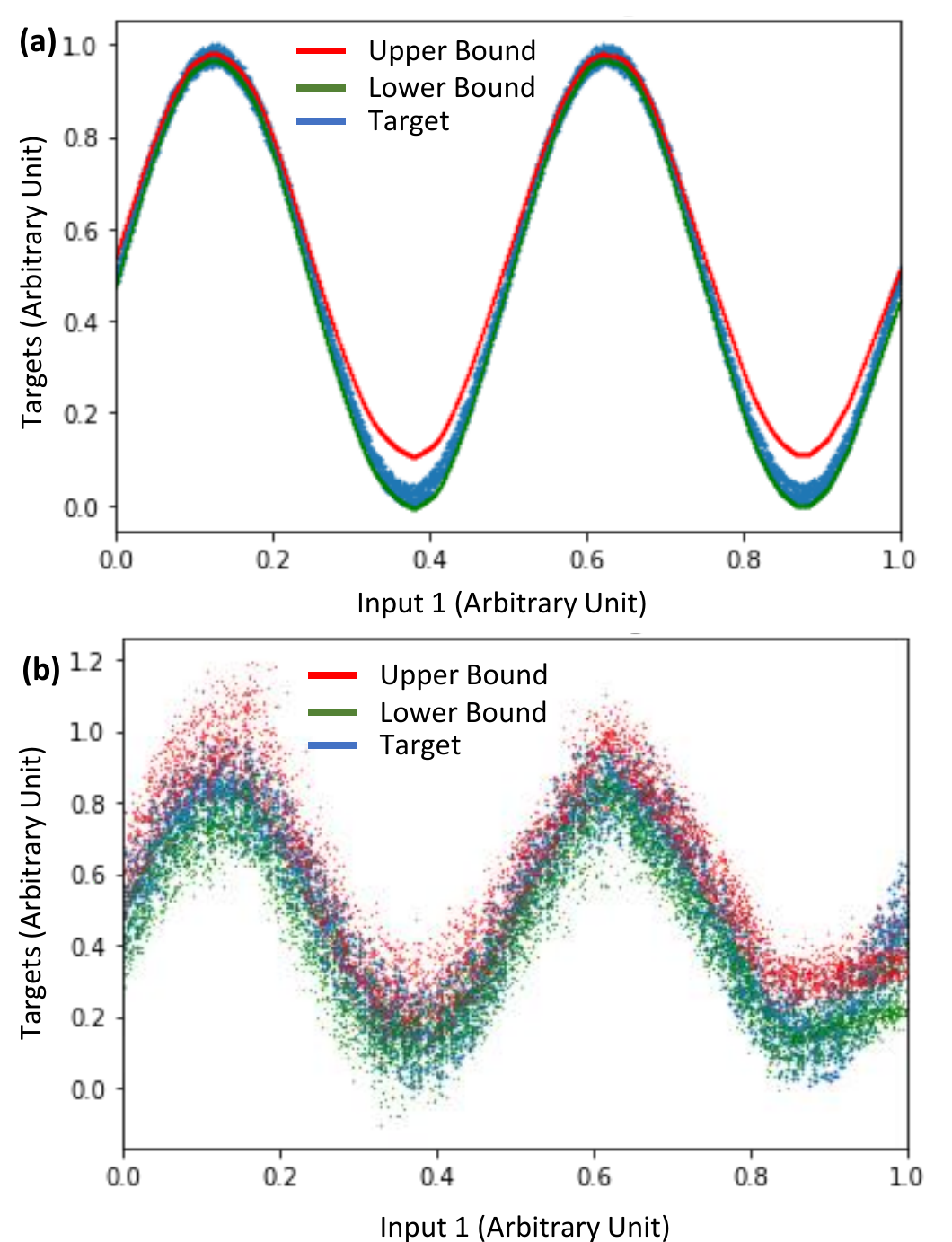}
  \caption{Targets with uncertainty bounds of 95\% coverage probability (a) on Dataset-1 (b) on Dataset-3 while applying the direct interval construction method with the optimal cost function. }
  \label{DataDirect}
\end{figure}

\fi

\subsection{Distribution Free Direct Intervals}
Khosravi et. al. first proposed the distribution-free direct interval method, known as the lower-upper bound estimation (LUBE) method \cite{khosravi2010lower}. Bayesian NNs have assumptions on the distribution of uncertainties. The direct interval construction method trains NNs directly with the help of a reward-based cost function.

Direct interval training in this paper consists of initial NN training with rough targets, followed by a final training. The direct NN consists of two hidden layers, each layer containing one thousand neurons. The initial training consists of one thousand epochs while the final training consists of five thousand epochs. Both traiing uses the Adam optimizer. The initial training has a learning rate of 5e-2 and the final training has a learnin rate of 5e-5. The equation for generating a rough target is as follows:
\begin{equation}
\label{Eq:Trough}
t_{j, rough} = t_j \pm \Delta t
\end{equation}
$t_{j, rough}$ contains two rough targets one is $t_j + \Delta t$ and another one is $t_j - \Delta t$. We kept $\Delta t = R/4$ based on observations. Future user of our script can adjust value of margin ($\Delta t$) based on the level of uncertainty on their dataset. 

\subsubsection{LUBE Method}
The LUBE method train NNs following a cost function, known as Coverage Width Criteria (CWC). The CWC is expressed as Eq. (\ref{CWC2}). 

The first segment of Table \ref{TABCost} presents the performance of the LUBE method on investigated Datasets. The LUBE cost function is designed to prevent lower coverage with a high penalty $\gamma  (PICP)$. Therefore, most of the LUBE NNs construct an interval that is wide enough to cover the required number of samples. As $\gamma  (PICP)$ becomes zero for PICP $> 1-\alpha$, the CWC becomes equal to PINAW for PICP $> 1-\alpha$. The term $1-\alpha$ is also known as the Prediction Interval Nominal Coverage (PINC).

The prediction interval normalized failure distance (PINAFD) is a parameter described in the optimal cost function subsection. Although PINAW generally becomes smaller for lower PINC, PINAFD may increase or decrease based on PINC. Most failure distances increase while intervals become narrower. However, more samples become uncovered for narrower PIs. As a result, the average failure distance often decreases. 

\subsubsection{Deviation from the Mid Interval Method}
L G Marin et al. proposed deviation from the mid-interval method \cite{marin2016prediction}. They consider a smooth PI failure penalty. Besides PINAW and PI failure penalties, they consider the nearness of mid-intervals from targets. Their cost function is as follows:
\begin{equation}%\label{Eq:cost2}
\underset{NN}{\text{min}} \ \ \beta_1 PINAW + \beta_2 ||e||^2 + exp[-\eta \big(PICP-(1-\alpha)\big)]
\end{equation}
where, $\beta_1$, and $\beta_2$ are hyper parameters,  $||e||$ is defined by the following equation:
\begin{equation}%\label{Eq:cost2}
||e|| = \sqrt{ \sum_{j=1}^n \Big|t_j- \frac{\overline{y}_j + \underline{y}_j}{2}\Big|^2 }.
\end{equation}
The second segment of Table \ref{TABCost} presents performances of prediction intervals developed by deviation from the mid-interval method. As the deviation from the mid-interval method has a slightly lower penalty for less coverage and more reward for narrowing PIs, PIs are slightly narrower. Also, PICP frequently becomes lower than PINC. Although the Table \ref{TABCost} presents average values of 100 iterations. PICP is lower than PINC on average in most of the combinations. 

\begin{table}
       
		\caption{PI Optimization Performance for Bayesian Neural Network-based methods.}
		\label{TABComp1}
		 \begin{adjustbox}{width=3.5in,center}
			\begin{tabular}{|c|c|c|c|c|c|c|c|c|}	\hline
				
			\multicolumn{9}{|c|}{Bayesian Method: Pyro \cite{blundell2015weight,Pyro}}\\ \hline
		    Data &$1-\alpha$    &$N_{Trial}$    &$\mu_{PINAW}$    &	$\mu_{PICP} $ 	& $\sigma_{PICP}$   & $\mu_{PINAFD}$ & $\mu_{CWC}$ & $\mu_{CWFDC}$ \\ \hline
			\multirow{ 3}{*}{\rotatebox[origin=c]{90}{\parbox[c]{1cm}{\centering Data-1}}}
			&0.95&5	    &1.72		&100\%   &0\%    &0 &1.72 &2.848\\ \cline{2-9}
			&0.90&5	    &1.41		&100\%   &0\%    &0 &1.41 &5.923\\ \cline{2-9}	
			&0.80&5	    &1.11		&100\%   &0\%    &0 &1.11 &19.16\\ \hline 
			
			\multirow{ 3}{*}{\rotatebox[origin=c]{90}{\parbox[c]{1cm}{\centering Data-3}}}
			&0.95&5	    &1.86		&100\%   &0\%    &0 &1.86 &2.988\\ \cline{2-9}
			&0.90&5	    &1.55		&100\%   &0\%    &0 &1.55 &6.063\\ \cline{2-9}	
			&0.80&5	    &1.31		&100\%   &0\%    &0 &1.31 &19.36\\ \hline 
			
			\multirow{ 3}{*}{\rotatebox[origin=c]{90}{\parbox[c]{1cm}{\centering Life Exp.}}}
			&0.95&5	    &2.02		&100\%   &0\%  &0 &2.02 &3.148\\ \cline{2-9}
			&0.90&5	    &1.69		&100\%   &0\%  &0 &1.69 &6.203\\ \cline{2-9}	
			&0.80&5	    &1.42		&100\%   &0\%  &0 &1.42 &19.47\\ \hline 
			
			\multirow{ 3}{*}{\rotatebox[origin=c]{90}{\parbox[c]{1cm}{\centering Elec. Demand}}}
			&0.95&5	    &1.21		&100\%   &0\%  &0 &1.21 &2.338\\ \cline{2-9}
			&0.90&5	    &1.01		&100\%   &0\%  &0 &1.01 &5.523\\ \cline{2-9}	
			&0.80&5	    &0.85		&100\%   &0\%  &0 &0.85 &18.9\\ \hline 
			
			\multirow{ 3}{*}{\rotatebox[origin=c]{90}{\parbox[c]{1cm}{\centering Wind Power}}}
			&0.95&5	    &1.38		&100\%   &0\%  &0 &1.38 &2.508\\ \cline{2-9}
			&0.90&5	    &1.15		&100\%   &0\%  &0 &1.15 &5.663\\ \cline{2-9}	
			&0.80&5	    &0.97		&100\%   &0\%  &0 &0.97 &19.02\\ \hline 
			\hline 
			
			\multicolumn{9}{|c|}{Bayesian Method: Pyro with Point Prediction NN}\\ \hline
		    Data &$1-\alpha$    &$N_{Trial}$    &$\mu_{PINAW}$    &	$\mu_{PICP}$ 	& $\sigma_{PICP}$   & $\mu_{PINAFD}$ & $\mu_{CWC}$ & $\mu_{CWFDC}$ \\ \hline
			\multirow{ 3}{*}{\rotatebox[origin=c]{90}{\parbox[c]{1cm}{\centering Data-1}}}
			&0.95&5	    &0.15		&100\%   &0\%&0 &\bf{0.15} &\bf{1.278}\\ \cline{2-9}
			&0.90&5	    &0.12		&100\%   &0\%&0 &\bf{0.12} &\bf{4.633}\\ \cline{2-9}	
			&0.80&5	    &0.11		&96.2\%   &3.2\%&0.01 &\bf{0.11} &\bf{11.672}\\ \hline 
			
			\multirow{ 3}{*}{\rotatebox[origin=c]{90}{\parbox[c]{1cm}{\centering Data-3}}}
			&0.95&5	    &0.32		&100\%   &0\%&0 &\bf{0.32} &\bf{1.448}\\ \cline{2-9}
			&0.90&5	    &0.27		&98.1\%   &1.6\%&0.012 &\bf{0.27} &\bf{3.17}\\ \cline{2-9}	
			&0.80&5	    &0.23		&93.1\%   &4.3\%&0.061 &\bf{0.23} &\bf{7.612}\\ \hline 
			
			\multirow{ 3}{*}{\rotatebox[origin=c]{90}{\parbox[c]{1cm}{\centering Life Exp.}}}
			&0.95&5	    &0.71		&100\%   &0\%&0 &\bf{0.71} &\bf{1.838}\\ \cline{2-9}
			&0.90&5	    &0.59		&95.7\%   &2.1\% &0.039 &\bf{0.59} &\bf{1.981}\\ \cline{2-9}	
			&0.80&5	    &0.50		&90.9\%   &3.8\%  &0.058 &\bf{0.50} &\bf{5.459}\\ \hline 
			
			\multirow{ 3}{*}{\rotatebox[origin=c]{90}{\parbox[c]{1cm}{\centering Elec. Demand}}}
			&0.95&5	    &0.24		&100\%   &0\%&0 &\bf{0.24} &\bf{1.368}\\ \cline{2-9}
			&0.90&5	    &0.20		&100\%   &0\%&0 &\bf{0.20} &\bf{4.713}\\ \cline{2-9}	
			&0.80&5	    &0.17		&93.5\%   &4.9\% &0.033 &\bf{0.17} &\bf{8.016}\\ \hline 
			
			\multirow{ 3}{*}{\rotatebox[origin=c]{90}{\parbox[c]{1cm}{\centering Wind Power}}}
			&0.95&5	    &0.62		&100\%   &0\%&0 &\bf{0.62} &\bf{1.748}\\ \cline{2-9}
			&0.90&5	    &0.52		&97.1\%   &1.1\%  &0.011 &\bf{0.52} &\bf{2.709}\\ \cline{2-9}	
			&0.80&5	    &0.44		&95.0\%   &1.8\%  &0.035 &\bf{0.44} &\bf{10.275}\\ \hline 
		    \hline 
			
			\multicolumn{9}{|c|}{Bayesian Method: TorchBNN Module \cite{lee2022graddiv}}\\ \hline
		    Data &$1-\alpha$    &$N_{Trial}$    &$\mu_{PINAW}$    &	$\mu_{PICP}$ 	& $\sigma_{PICP}$   & $\mu_{PINAFD}$ & $\mu_{CWC}$ & $\mu_{CWFDC}$ \\ \hline
			\multirow{ 3}{*}{\rotatebox[origin=c]{90}{\parbox[c]{1cm}{\centering Data-1}}}
			&0.95&5	    &0.05		&81.3\%&4.2\%   &0.006 &47.244 &9.786\\ \cline{2-9}
			&0.90&5	    &0.04		&76.7\%&4.5\%   &0.012 &30.951 &9.574\\ \cline{2-9}	
			&0.80&5	    &0.03		&62.5\%&4.9\%   &0.011 &189.351 &17.154\\ \hline 
			
			\multirow{ 3}{*}{\rotatebox[origin=c]{90}{\parbox[c]{1cm}{\centering Data-3}}}
			&0.95&5	    &0.14		&69.1\%&6.1\%   &0.010 &5.891e4 &34.341\\ \cline{2-9}
			&0.90&5	    &0.11		&50.3\%&6.8\%   &0.015 &4.593e7 &80.927\\ \cline{2-9}	
			&0.80&5	    &0.10		&41.9\%&5.9\%   &0.019 &1.876e3 &76.560\\ \hline 
			
			\multirow{ 3}{*}{\rotatebox[origin=c]{90}{\parbox[c]{1cm}{\centering Life Exp.}}}
			&0.95&5	    &0.43		&67.3\%&6.2\%    &0.052 &4.45e5 &39.542\\ \cline{2-9}
			&0.90&5	    &0.39		&56.1\%&6.9\%    &0.050 &8.961e6 &59.608\\ \cline{2-9}	
			&0.80&5	    &0.38		&51.9\%&7.2\%    &0.055 &4.804e5 &42.776\\ \hline 
			
			\multirow{ 3}{*}{\rotatebox[origin=c]{90}{\parbox[c]{1cm}{\centering Elec. Demand}}}
			&0.95&5	    &0.15		&91.0\%&1.1\%    &0.011 &1.258 &1.064\\ \cline{2-9}
			&0.90&5	    &0.14		&82.6\%&3.1\%    &0.012 &5.803 &3.273\\ \cline{2-9}	
			&0.80&5	    &0.13		&71.8\%&3.7\%    &0.009 &7.974 &4.371\\ \hline 
			
			\multirow{ 3}{*}{\rotatebox[origin=c]{90}{\parbox[c]{1cm}{\centering Wind Power}}}
			&0.95&5	    &0.21		&84.3\%&3.0\%    &0.018 &44.437 &6.223\\ \cline{2-9}
			&0.90&5	    &0.19		&69.1\%&3.9\%    &0.020 &6.563e3 &23.108\\ \cline{2-9}	
			&0.80&5	    &0.18		&62.7\%&4.3\%    &0.022 &1.028e3 &16.947\\ \hline
			
			\end{tabular} 
			\end{adjustbox}

\end{table}

\subsubsection{Optimal Cost Function Method}
The optimal cost function is developed by combining the philosophies of several recently proposed cost functions \cite{kabir2021optimal}. The cost function is as follows:
\begin{equation}\label{Eq:CWFDC}
\underset{NN}{\text{min}} \ \ PINAW +\rho . PINAFD + \beta . (1-\alpha+\delta -PICP)^2
\end{equation}
where, $\rho$ is the failure distance resistance parameter. $\rho$ becomes equal to one when providing an equal concentration to PINAW and PINAFD. $\beta$ is the PI failure penalty factor, $\alpha$ is the target non-coverage probability, and $\delta$ is the PICP margin. Usually, $\delta = \alpha/20$ that brings PICP $>$ PINC statistically. The prediction interval normalized average failure distance (PINAFD) is expressed as follows:
\begin{equation}
\label{Eq:PINAFD}
PINAFD = \frac{\sum_{j=1}^n (1-c_j) \times min(|t_j-\overline{y}_j|,|\underline{y}_j-t_j|)}{R \times \Big\{\sum_{j=1}^n (1-c_j)\Big\} +\epsilon }.
\end{equation}
where, 
\begin{equation*}
c_j = 	\begin{cases}
		1, t_j \in [\underline{y}_j, \overline{y}_j] \\
		0, t_j \not \in [\underline{y}_j, \overline{y}_j],
		\end{cases}
\end{equation*}
where, $t_j$ is the target for $j^{th}$ sample. $R$ is the range of targets, $\epsilon \approx (1e-10)$ is a very small number added to avoid the divided by zero error, $\overline{y}_j$ and $\underline{y}_j$ are the predicted upper bound and lower bound respectively for $j^{th}$ sample. This cost function is also known as the coverage width failure distance criteria (CWFDC).
A publicly available script is also available with execution details of optimal cost-function-based training on the GitHub repository of this paper. The script contains an extra bound swap penalty parameter to ensure that there is no bound swap. The script also performs an initial training with rough targets before the cost function based training. Rough targets are generated by adding a margin with the original target. Initial training with rough targets increases convergence of NN training and reduces training time significantly \cite{kabir2020neural}. 

Fig. \ref{DataUB} (c)-(d) present targets with uncertainty bounds of 95\% coverage probability while applying the direct interval construction method with the optimal cost function \cite{kabir2021optimal}. Fig. \ref{DataUB}(c) and Fig. \ref{DataUB}(d) presents targets with intervals for Dataset-1 and Dataset-3 respectively. The third segment of Table \ref{TABCost} presents the performance of the optimal cost function on investigated datasets. The optimal cost function provides superior performance on average compared to Bayesian and other cost-function-based training methods.

\begin{table}
       
		\caption{PI Optimization Performance for Cost function-based Direct Interval Construction methods.}
		\label{TABCost}
		 \begin{adjustbox}{width=3.5in,center}
			\begin{tabular}{|c|c|c|c|c|c|c|c|c|}	\hline
				
			\multicolumn{9}{|c|}{LUBE Method \cite{khosravi2010lower}}\\ \hline
		    Data &$1-\alpha$    &$N_{Trial}$    &$\mu_{PINAW}$    &	$\mu_{PICP}$ 	& $\sigma_{PICP}$   & $\mu_{PINAFD}$ & $\mu_{CWC}$ & $\mu_{CWFDC}$ \\ \hline
			\multirow{ 3}{*}{\rotatebox[origin=c]{90}{\parbox[c]{1cm}{\centering Data-1}}}
			&0.95&100	    &0.072		&96.2\%  &2.2\%    &0.01 &0.072 &0.127\\ \cline{2-9}
			&0.90&100	    &0.070		&90.9\%   &3.1\%    &0.02 &0.070 &0.098\\ \cline{2-9}	
			&0.80&100	    &0.069		&82.1\%   &5.3\%    &0.01 &0.069 &0.140\\ \hline 
			
			\multirow{ 3}{*}{\rotatebox[origin=c]{90}{\parbox[c]{1cm}{\centering Data-3}}}
			&0.95&100	    &0.177		&95.5\%   &1.6\%    &0.02 &\bf{\emph{0.177}} &\bf{\emph{0.200}}\\ \cline{2-9}
			&0.90&100	    &0.178		&90.7\%   &2.3\%    &0.03 &0.178 &0.210\\ \cline{2-9}	
			&0.80&100	    &0.177		&81.6\%  &4.2\%    &0.01 &0.177 &\bf{\emph{0.205}}\\ \hline 
			
			\multirow{ 3}{*}{\rotatebox[origin=c]{90}{\parbox[c]{1cm}{\centering Life Exp.}}}
			&0.95&100	    &0.475		&95.9\%   &2.1\%    &0.09 &\bf{0.475} &\bf{0.586}\\ \cline{2-9}
			&0.90&100	    &0.462		&89.9\%   &3.3\%    &0.10 &0.948 &0.580\\ \cline{2-9}	
			&0.80&100	    &0.431		&83.3\%   &5.2\%    &0.08 &\bf{\emph{0.431}} &0.776\\ \hline 
			
			\multirow{ 3}{*}{\rotatebox[origin=c]{90}{\parbox[c]{1cm}{\centering Elec. Demand}}}
			&0.95&100	    &0.171		&95.3\%   &1.0\%    &0.02 &\bf{0.171} &0.191\\ \cline{2-9}
			&0.90&100	    &0.163		&90.7\%   &1.9\%    &0.03 &\bf{\emph{0.163}} &\bf{0.195}\\ \cline{2-9}	
			&0.80&100	    &0.160		&81.2\%   &2.1\%    &0.02 &\bf{\emph{0.160}} &\bf{\emph{0.182}}\\ \hline 
			
			\multirow{ 3}{*}{\rotatebox[origin=c]{90}{\parbox[c]{1cm}{\centering Wind Power}}}
			&0.95&100	    &0.331		&96.0\%   &1.6\%    &0.06 &0.331 &0.419\\ \cline{2-9}
			&0.90&100	    &0.312		&90.6\%   &2.9\%    &0.05 &0.312 &0.363\\ \cline{2-9}	
			&0.80&100	    &0.300		&80.0\%   &6.5\%    &0.05 &\bf{0.300}	&0.400\\ \hline 
			\hline 
			
			\multicolumn{9}{|c|}{Deviation from Mid Interval method \cite{marin2016prediction}}\\ \hline
		    Data &$1-\alpha$    &$N_{Trial}$    &$\mu_{PINAW}$    &	$\mu_{PICP}$ 	& $\sigma_{PICP}$   & $\mu_{PINAFD}$ & $\mu_{CWC}$ & $\mu_{CWFDC}$ \\ \hline
			\multirow{ 3}{*}{\rotatebox[origin=c]{90}{\parbox[c]{1cm}{\centering Data-1}}}
			&0.95&100	    &0.071		&95.0\%   &1.9\%    &0.015 &0.071 &0.089\\ \cline{2-9}
			&0.90&100	    &0.069		&89.8\%   &2.6\%    &0.014 &0.145 &0.108\\ \cline{2-9}	
			&0.80&100	    &0.068		&79.5\%   &4.3\%    &0.011 &0.155 &0.192\\ \hline 
			
			\multirow{ 3}{*}{\rotatebox[origin=c]{90}{\parbox[c]{1cm}{\centering Data-3}}}
			&0.95&100	    &0.182		&94.8\%   &1.3\%    &0.031 &0.383 &0.223\\ \cline{2-9}
			&0.90&100	    &0.180		&91.6\%   &2.1\%    &0.033 &0.180 &0.274\\ \cline{2-9}	
			&0.80&100	    &0.176		&80.3\%   &3.3\%   &0.027 &0.176 &0.228\\ \hline 
			
			\multirow{ 3}{*}{\rotatebox[origin=c]{90}{\parbox[c]{1cm}{\centering Life Exp.}}}
			&0.95&100	    &0.470		&94.2\%   &1.8\%    &0.085 &1.171 &0.610\\ \cline{2-9}
			&0.90&100	    &0.463		&89.7\%   &3.3\%    &0.113 &1.001 &0.608\\ \cline{2-9}	
			&0.80&100	    &0.429		&78.1\%   &5.0\%    &0.101 &1.538 &0.951\\ \hline 
			
			\multirow{ 3}{*}{\rotatebox[origin=c]{90}{\parbox[c]{1cm}{\centering Elec. Demand}}}
			&0.95&100	    &0.175		&95.1\%   &0.9\%    &0.029 &0.175 &0.205\\ \cline{2-9}
			&0.90&100	    &0.168		&90.3\%   &1.6\%    &0.025 &0.168 &\bf{0.195}\\ \cline{2-9}	
			&0.80&100	    &0.167		&80.7\%   &1.9\%    &0.021 &0.167 &0.193\\ \hline 
			
			\multirow{ 3}{*}{\rotatebox[origin=c]{90}{\parbox[c]{1cm}{\centering Wind Power}}}
			&0.95&100	    &0.331		&94.8\%   &1.5\%    &0.046 &0.697 &\bf{0.387}\\ \cline{2-9}
			&0.90&100	    &0.312		&89.6\%   &2.7\%    &0.039 &0.693 &0.392\\ \cline{2-9}	
			&0.80&100	    &0.300		&78.2\%   &4.2\%    &0.038 &1.038 &0.730\\ \hline 
			\hline 
			
			\multicolumn{9}{|c|}{Optimal Cost Function \cite{kabir2021optimal}}\\ \hline
		    Data &$1-\alpha$    &$N_{Trial}$    &$\mu_{PINAW}$    &	$\mu_{PICP}$ 	& $\sigma_{PICP}$   & $\mu_{PINAFD}$ & $\mu_{CWC}$ & $\mu_{CWFDC}$ \\ \hline
			\multirow{ 3}{*}{\rotatebox[origin=c]{90}{\parbox[c]{1cm}{\centering Data-1}}}
			&0.95&100	    &0.069		&95.1\%   &1.1\%    &0.005 &\bf{0.069} &\bf{0.075}\\ \cline{2-9}
			&0.90&100	    &0.068		&90.5\%   &1.8\%    &0.008 &\bf{0.068} &\bf{0.076}\\ \cline{2-9}	
			&0.80&100	    &0.067		&80.9\%   &2.7\%    &0.006 &\bf{0.067} &\bf{\emph{0.074}}\\ \hline 
			
			\multirow{ 3}{*}{\rotatebox[origin=c]{90}{\parbox[c]{1cm}{\centering Data-3}}}
			&0.95&100	    &0.179		&95.0\%   &0.8\%    &0.020 &0.179 &0.202\\ \cline{2-9}
			&0.90&100	    &0.175		&90.3\%   &1.4\%    &0.013 &\bf{\emph{0.175}} &\bf{\emph{0.190}}\\ \cline{2-9}	
			&0.80&100	    &0.172		&80.2\%   &2.6\%    &0.017 &\bf{\emph{0.172}} &0.221\\ \hline 
			
			\multirow{ 3}{*}{\rotatebox[origin=c]{90}{\parbox[c]{1cm}{\centering Life Exp.}}}
			&0.95&100	    &0.466		&93.8\%   &1.4\%    &0.078 &1.315 &0.649\\ \cline{2-9}
			&0.90&100	    &0.461		&90.0\%   &2.6\%    &0.081 &\bf{0.461} &\bf{0.555}\\ \cline{2-9}	
			&0.80&100	    &0.440		&79.9\%   &3.8\%    &0.084 &0.903 &\bf{0.585}\\ \hline 
			
			\multirow{ 3}{*}{\rotatebox[origin=c]{90}{\parbox[c]{1cm}{\centering Elec. Demand}}}
			&0.95&100	    &0.173		&95.1\%   &0.8\%    &0.016 &0.173 &\bf{\emph{0.190}}\\ \cline{2-9}
			&0.90&100	    &0.167		&90.0\%   &1.5\%    &0.018 &0.167 &0.198\\ \cline{2-9}	
			&0.80&100	    &0.162		&80.2\%   &1.8\%    &0.017 &0.162 &0.211\\ \hline 
			
			\multirow{ 3}{*}{\rotatebox[origin=c]{90}{\parbox[c]{1cm}{\centering Wind Power}}}
			&0.95&100	    &0.327		&96.1\%   &1.1\%    &0.024 &\bf{0.327} &\bf{0.387}\\ \cline{2-9}
			&0.90&100	    &0.310		&90.9\%   &2.4\%    &0.019 &\bf{\emph{0.310}} &\bf{0.337}\\ \cline{2-9}	
			&0.80&100	    &0.307		&81.6\%   &2.9\%    &0.020 &0.307 &\bf{\emph{0.345}}\\ \hline 
			
			\end{tabular} 
			\end{adjustbox}

\end{table}	

\subsection{Proposed Method}
The direct interval construction method with optimal cost function provides a sharp interval with nearly accurate coverage. However, there is still debate on the selection of appropriate cost functions. Some researchers think that only narrowing PIs should be the goal. Some other researchers think about getting a small failure distance. Some other researchers are trying to bring the mid-interval near the target. Another approach is training NNs for uncertainty-bound values. However, cost-function-based training NNs for uncertainty-bound values brings poor NNs. Neural networks with such training often predict a constant which brings statistical average UB instead of a heteroscedastic uncertainty bound.
Therefore, we propose a similarity-based method to find UB and train NNs with obtained bounds. 

The initial point prediction NN training consists of six hundred epochs. The error prediction NN also requires the same number of epochs. Both NNs contain two hidden layers and each layer contains five hundred neurons. The learning rate is 0.05 for both situations. The sample density NN contains two hidden layers and each layer contains five hundred neurons. The learning rate is also 0.05 for the sample density NN training. The bound correction NN also has the same model and training attributes. The final NN size and training epoch varies based on the complexity of the data. Training scripts are available at the GitHub repository. 

Fig. \ref{DataUB} presents targets with uncertainty upper bounds of intervals of 90\% coverage probability while applying the proposed similarity, sensitivity, and error sensitivity-based method.  The method constructs an uncertainty bound of a certain cumulative probability value. 
Fig. \ref{DataUB}(a) presents targets of Dataset-1 with an uncertainty upper bound of 95\% cumulative probability. Fig. \ref{DataUB}(b) presents targets of Dataset-3 with an uncertainty upper bound of 95\% cumulative probability. This figure shows denormalized values. Therefore, the ranges in both axis are different from the normalized range (0 to 1). 
Table \ref{TABSim} presents both the proposed method and our recently proposed similarity and sensitivity-based method \cite{kabir2020uncertainty}. Our recently proposed paper \cite{kabir2020uncertainty} considers similarity and prediction sensitivity. The method proposed in this paper considers similarity, prediction sensitivity, and error sensitivity. Earlier in this paper, Fig. \ref{Input_uncertainty} explained the effect of discarding one parameter which only contributes to uncertainty.

\begin{table}
       
		\caption{PI Optimization Performance for Similarity-based Approaches.}
		\label{TABSim}
		 \begin{adjustbox}{width=3.5in,center}
			\begin{tabular}{|c|c|c|c|c|c|c|c|c|}	\hline
				
			\multicolumn{9}{|c|}{UB from Similarity and Sensitivity \cite{kabir2020uncertainty}}\\ \hline
		    Data &$1-\alpha$    &$N_{Trial}$    &$\mu_{PINAW}$    &	$\mu_{PICP}$ 	& $\sigma_{PICP}$   & $\mu_{PINAFD}$ & $\mu_{CWC}$ & $\mu_{CWFDC}$ \\ \hline
			\multirow{ 3}{*}{\rotatebox[origin=c]{90}{\parbox[c]{1cm}{\centering Data-1}}}
			&0.95&100	    &0.063		&95.0\%   &1.9\%    &0.004 &0.063 &0.070\\ \cline{2-9}
			&0.90&100	    &0.062		&90.1\%   &2.1\%    &0.005 &0.062 &0.075\\ \cline{2-9}	
			&0.80&100	    &0.060		&79.8\%   &2.6\%    &0.004 &0.126 &0.136\\ \hline 
			
			\multirow{ 3}{*}{\rotatebox[origin=c]{90}{\parbox[c]{1cm}{\centering Data-3}}}
			&0.95&100	    &0.191		&95.0\%   &0.5\%    &0.019 &0.191 &0.213\\ \cline{2-9}
			&0.90&100	    &0.185		&90.2\%   &0.8\%   &0.016 &\bf{0.185} &\bf{0.206}\\ \cline{2-9}	
			&0.80&100	    &0.180		&80.3\%   &1.8\%    &0.012 &0.180 &\bf{0.217}\\ \hline  
			
			\multirow{ 3}{*}{\rotatebox[origin=c]{90}{\parbox[c]{1cm}{\centering Life Exp.}}}
			&0.95&100	    &0.461		&94.8\%   &1.7\%    &0.081 &0.970 &0.552\\ \cline{2-9}
			&0.90&100	    &0.457		&90.9\%  &2.3\%    &0.077 &0.457 &0.542\\ \cline{2-9}	
			&0.80&100	    &0.451		&80.3\%   &4.2\%    &0.074 &0.451 &\bf{\emph{0.550}}\\ \hline 
			
			\multirow{ 3}{*}{\rotatebox[origin=c]{90}{\parbox[c]{1cm}{\centering Elec. Demand}}}
			&0.95&100	    &0.171		&95.0\%   &0.7\%    &0.019 &0.171 &0.193\\ \cline{2-9}
			&0.90&100	    &0.167		&90.1\%  &1.6\%    &0.017 &\bf{0.167} &\bf{\emph{0.192}}\\ \cline{2-9}	
			&0.80&100	    &0.164		&80.2\%   &2.2\%    &0.016 &0.164 &\bf{0.212}\\ \hline 
			
			\multirow{ 3}{*}{\rotatebox[origin=c]{90}{\parbox[c]{1cm}{\centering Wind Power}}}
			&0.95&100	    &0.327		&95.2\%   &0.9\%    &0.024 &0.327 &0.351\\ \cline{2-9}
			&0.90&100	    &0.311		&90.3\%   &1.3\%    &0.019 &\bf{0.311} &\bf{\emph{0.332}}\\ \cline{2-9}	
			&0.80&100	    &0.308		&80.1\%  &2.0\%    &0.017 &0.308 &0.366\\ \hline \hline

			\multicolumn{9}{|c|}{UB from Similarity, Sensitivity, and Error Sensitivity}\\ \hline
		    Data &$1-\alpha$    &$N_{Trial}$    &$\mu_{PINAW}$    &	$\mu_{PICP}$ 	& $\sigma_{PICP}$   & $\mu_{PINAFD}$ & $\mu_{CWC}$ & $\mu_{CWFDC}$ \\ \hline
			\multirow{ 3}{*}{\rotatebox[origin=c]{90}{\parbox[c]{1cm}{\centering Data-1}}}
			&0.95&100	    &0.062		&95.0\%   &1.8\%    &0.004 &\bf{\emph{0.062}} &\bf{\emph{0.069}}\\ \cline{2-9}
			&0.90&100	    &0.061		&90.1\%   &2.0\%    &0.004 &\bf{\emph{0.061}} &\bf{\emph{0.073}}\\ \cline{2-9}	
			&0.80&100	    &0.059		&80.2\%   &2.2\%    &0.003 &\bf{\emph{0.059}} &\bf{0.094}\\ \hline 
			
			\multirow{ 3}{*}{\rotatebox[origin=c]{90}{\parbox[c]{1cm}{\centering Data-3}}}
			&0.95&100	    &0.190		&95.0\%   &0.6\%    &0.018 &\bf{0.190} &\bf{0.211}\\ \cline{2-9}
			&0.90&100	    &0.185		&90.0\%   &0.7\%    &0.015 &\bf{0.185} &0.213\\ \cline{2-9}	
			&0.80&100	    &0.178		&80.2\%   &1.3\%    &0.013 &\bf{0.178} &0.223\\ \hline  
			
			\multirow{ 3}{*}{\rotatebox[origin=c]{90}{\parbox[c]{1cm}{\centering Life Exp.}}}
			&0.95&100	    &0.455		&95.1\%   &1.7\%    &0.083 &\bf{\emph{0.455}} &\bf{\emph{0.539}}\\ \cline{2-9}
			&0.90&100	    &0.453		&90.1\%   &2.2\%    &0.079 &\bf{\emph{0.453}} &\bf{\emph{0.540}}\\ \cline{2-9}	
			&0.80&100	    &0.450		&80.2\%   &2.9\%    &0.077 &\bf{0.450} &0.559\\ \hline 
			
			\multirow{ 3}{*}{\rotatebox[origin=c]{90}{\parbox[c]{1cm}{\centering Elec. Demand}}}
			&0.95&100	    &0.169		&95.0\%   &0.7\%    &0.019 &\bf{\emph{0.169}} &\bf{0.191}\\ \cline{2-9}
			&0.90&100	    &0.165		&89.9\%   &1.5\%    &0.018 &0.338 &0.201\\ \cline{2-9}	
			&0.80&100	    &0.162		&80.0\%   &2.0\%    &0.016 &\bf{0.162} &0.228\\ \hline 
			
			\multirow{ 3}{*}{\rotatebox[origin=c]{90}{\parbox[c]{1cm}{\centering Wind Power}}}
			&0.95&100	    &0.320		&95.1\%   &1.0\%    &0.023 &\bf{\emph{0.320}} &\bf{\emph{0.344}}\\ \cline{2-9}
			&0.90&100	    &0.297		&89.8\%   &1.2\%    &0.020 &0.625 &0.342\\ \cline{2-9}	
			&0.80&100	    &0.289		&80.0\%   &2.0\%    &0.017 &\bf{\emph{0.289}} &\bf{0.356}\\ \hline 	
			
			\end{tabular} 
			\end{adjustbox}

\end{table}

The proposed method statistically outperforms other investigated methods. All five datasets are investigated for three different nominal coverage probabilities. We compute CWC and CWFDC values for each combination. We present optimal values among the same type of training with bold-faced numbers. Among BNN methods Pyro BNN with point prediction support provides optimal results in all situations. Therefore, all values of cost functions are bold-faced in this segment. We present the optimum value of all methods with both bold and italic numbers.  

Among cost-function-based methods, the optimal cost function method provides superior performance in Dataset-1 and wind power data. LUBE and optimal cost function methods perform equally for Dataset-3 and life expectancy data. The LUBE method provides superior performance on the electricity demand dataset. Among similarity-based methods, both methods perform equally on the electricity demand dataset. The proposed method provides superior performance on the other four datasets.

The proposed method provides overall superior performance on Dataset-1, Life expectancy data, and Windpower generation data. 
The optimal cost function method provides overall superior performance on Dataset-3. The LUBE method provides overall superior performance on the Electricity demand dataset.

\section*{Acknowledgment}
This research was partially supported by the Australian Research Council's Discovery Projects funding scheme (project DP190102181).

\section{Conclusion} \label{secconc}
In this paper, we have proposed a similarity-based uncertainty quantification neural network training method. According to our \hlx{observation}, cost-function-based methods and similarity-based methods outperforms Bayesian methods. Bayesian methods are formulated based on strong assumptions and often fail to provide \hlx{optimal} coverage. Cost-function-based methods provide superior results in several investigated datasets. Similarity-based methods are better in other datasets. The proposed \hlx{method has provided} statistically superior performances in three out of five datasets. Besides the statistical numeric improvement on the quality of prediction interval, the \hlx{proposed process} has the provision to plot similar samples and to observe sample density near a sample. The plot of the distribution of similar samples can potentially convey a more detailed idea of the distribution of uncertainties. We have proposed the method with publicly available codes and with execution details as Jupyter Notebook files. Several scripts of Bayesian NN and direct interval construction are also provided with execution details to help future researchers. We have also explained several phenomena and reasons for performing specific actions with relevant sketches and diagrams. Future researchers can potentially apply the proposed method \hlx{to} new datasets. Future researchers can also propose a more robust method and compare results with the help of shared scripts.

\hlx{Although trained models in the proposed method statistically provide superior performance, the training is computationally extensive. We reduce the execution time by training NN for the prediction. Future researchers may apply a less computation extensive data selection technique to find sensitivity-aware similar samples. All the investigated methods, including the proposed method exhibit poor performance when the number of training samples is small. With a small number of training samples, prediction intervals become wide and fail to cover the target more frequently.}

\bibliographystyle{Citation_Styles/model6-num-names.bst}

\bibliography{References}

\begin{thebibliography}{49}
\providecommand{\natexlab}[1]{#1}
\providecommand{\url}[1]{\texttt{#1}}
\providecommand{\urlprefix}{URL }
\expandafter\ifx\csname urlstyle\endcsname\relax
  \providecommand{\doi}[1]{doi:\discretionary{}{}{}#1}\else
  \providecommand{\doi}{doi:\discretionary{}{}{}\begingroup
  \urlstyle{rm}\Url}\fi
\providecommand{\eprint}[2][]{\url{#2}}
\providecommand{\BIBand}{and}
\providecommand{\bibinfo}[2]{#2}
\ifx\xfnm\undefined \def\xfnm[#1]{\unskip,\space#1}\fi
\makeatletter\def\@biblabel#1{#1.}\makeatother
%Type = Article
\bibitem[{Theisen et~al.(2021)Theisen, Wang, Varshney, Xiong and
  Socher}]{theisen2021evaluating}
\bibinfo{author}{Theisen\xfnm[ R.]}, \bibinfo{author}{Wang\xfnm[ H.]},
  \bibinfo{author}{Varshney\xfnm[ L.R.]}, \bibinfo{author}{Xiong\xfnm[ C.]},
  \bibinfo{author}{Socher\xfnm[ R.]}.
\newblock \bibinfo{title}{Evaluating state-of-the-art classification models
  against bayes optimality}.
\newblock \emph{\bibinfo{journal}{Advances in Neural Information Processing
  Systems}}
  \bibinfo{year}{2021};\bibinfo{volume}{34}:\bibinfo{pages}{9367--9377}.
%Type = Article
\bibitem[{Fan et~al.(2021)Fan, Wang, Guo, Zhu, Yan, Wang and
  Yu}]{fan2021sparse}
\bibinfo{author}{Fan\xfnm[ F.L.]}, \bibinfo{author}{Wang\xfnm[ D.]},
  \bibinfo{author}{Guo\xfnm[ H.]}, \bibinfo{author}{Zhu\xfnm[ Q.]},
  \bibinfo{author}{Yan\xfnm[ P.]}, \bibinfo{author}{Wang\xfnm[ G.]},
  \bibinfo{author}{Yu\xfnm[ H.]}.
\newblock \bibinfo{title}{On a sparse shortcut topology of artificial neural
  networks}.
\newblock \emph{\bibinfo{journal}{IEEE Transactions on Artificial
  Intelligence}}
  \bibinfo{year}{2021};\bibinfo{volume}{3}(\bibinfo{number}{4}):\bibinfo{pages}{595--608}.
%Type = Article
\bibitem[{Koochali et~al.(2021)Koochali, Dengel and Ahmed}]{koochali2021if}
\bibinfo{author}{Koochali\xfnm[ A.]}, \bibinfo{author}{Dengel\xfnm[ A.]},
  \bibinfo{author}{Ahmed\xfnm[ S.]}.
\newblock \bibinfo{title}{If you like it, gan it—probabilistic multivariate
  times series forecast with gan}.
\newblock \emph{\bibinfo{journal}{Engineering Proceedings}}
  \bibinfo{year}{2021};\bibinfo{volume}{5}(\bibinfo{number}{1}):\bibinfo{pages}{40}.
%Type = Inproceedings
\bibitem[{Lakshminarayanan et~al.(2017)Lakshminarayanan, Pritzel and
  Blundell}]{lakshminarayanan2017simple}
\bibinfo{author}{Lakshminarayanan\xfnm[ B.]}, \bibinfo{author}{Pritzel\xfnm[
  A.]}, \bibinfo{author}{Blundell\xfnm[ C.]}.
\newblock \bibinfo{title}{Simple and scalable predictive uncertainty estimation
  using deep ensembles}.
\newblock In: \emph{\bibinfo{booktitle}{Advances in neural information
  processing systems}}. \bibinfo{year}{2017}:\unskip
  \bibinfo{pages}{6402--6413}.
%Type = Article
\bibitem[{Morala et~al.(2021)Morala, Cifuentes, Lillo and
  Ucar}]{morala2021towards}
\bibinfo{author}{Morala\xfnm[ P.]}, \bibinfo{author}{Cifuentes\xfnm[ J.A.]},
  \bibinfo{author}{Lillo\xfnm[ R.E.]}, \bibinfo{author}{Ucar\xfnm[ I.]}.
\newblock \bibinfo{title}{Towards a mathematical framework to inform neural
  network modelling via polynomial regression}.
\newblock \emph{\bibinfo{journal}{Neural Networks}}
  \bibinfo{year}{2021};\bibinfo{volume}{142}:\bibinfo{pages}{57--72}.
%Type = Article
\bibitem[{Mir et~al.(2021)Mir, Kabir, Nasirzadeh and Khosravi}]{mir2021neural}
\bibinfo{author}{Mir\xfnm[ M.]}, \bibinfo{author}{Kabir\xfnm[ H.D.]},
  \bibinfo{author}{Nasirzadeh\xfnm[ F.]}, \bibinfo{author}{Khosravi\xfnm[ A.]}.
\newblock \bibinfo{title}{Neural network-based interval forecasting of
  construction material prices}.
\newblock \emph{\bibinfo{journal}{Journal of Building Engineering}}
  \bibinfo{year}{2021};\bibinfo{volume}{39}:\bibinfo{pages}{102288}.
%Type = Article
\bibitem[{Kavousi-Fard et~al.(2020)Kavousi-Fard, Dabbaghjamanesh, Jin, Su and
  Roustaei}]{kavousi2020evolutionary}
\bibinfo{author}{Kavousi-Fard\xfnm[ A.]},
  \bibinfo{author}{Dabbaghjamanesh\xfnm[ M.]}, \bibinfo{author}{Jin\xfnm[ T.]},
  \bibinfo{author}{Su\xfnm[ W.]}, \bibinfo{author}{Roustaei\xfnm[ M.]}.
\newblock \bibinfo{title}{An evolutionary deep learning-based anomaly detection
  model for securing vehicles}.
\newblock \emph{\bibinfo{journal}{IEEE Transactions on Intelligent
  Transportation Systems}}
  \bibinfo{year}{2020};\bibinfo{volume}{22}(\bibinfo{number}{7}):\bibinfo{pages}{4478--4486}.
%Type = Article
\bibitem[{Dong et~al.(2021)Dong, Gong and Zhu}]{dong2021optimal}
\bibinfo{author}{Dong\xfnm[ H.]}, \bibinfo{author}{Gong\xfnm[ Q.]},
  \bibinfo{author}{Zhu\xfnm[ M.]}.
\newblock \bibinfo{title}{Optimal energy management of automated grids
  considering the social and technical objectives with electric vehicles}.
\newblock \emph{\bibinfo{journal}{International Journal of Electrical Power \&
  Energy Systems}}
  \bibinfo{year}{2021};\bibinfo{volume}{130}:\bibinfo{pages}{106910}.
%Type = Article
\bibitem[{Zare et~al.(2022)Zare, Shoeibi, Bajestani, Moridian, Alizadehsani,
  Hallaji and Khosravi}]{zare2022accurate}
\bibinfo{author}{Zare\xfnm[ A.]}, \bibinfo{author}{Shoeibi\xfnm[ A.]},
  \bibinfo{author}{Bajestani\xfnm[ N.S.]}, \bibinfo{author}{Moridian\xfnm[
  P.]}, \bibinfo{author}{Alizadehsani\xfnm[ R.]},
  \bibinfo{author}{Hallaji\xfnm[ M.]}, \bibinfo{author}{Khosravi\xfnm[ A.]}.
\newblock \bibinfo{title}{Accurate prediction using triangular type-2 fuzzy
  linear regression: Simplifying complex t2f calculations}.
\newblock \emph{\bibinfo{journal}{IEEE Systems, Man, and Cybernetics Magazine}}
  \bibinfo{year}{2022};\bibinfo{volume}{8}(\bibinfo{number}{3}):\bibinfo{pages}{51--60}.
%Type = Article
\bibitem[{Mobtahej et~al.(2021)Mobtahej, Esapour, Tajalli and
  Mohammadi}]{mobtahej2021effective}
\bibinfo{author}{Mobtahej\xfnm[ M.]}, \bibinfo{author}{Esapour\xfnm[ K.]},
  \bibinfo{author}{Tajalli\xfnm[ S.Z.]}, \bibinfo{author}{Mohammadi\xfnm[ M.]}.
\newblock \bibinfo{title}{Effective demand response and gans for optimal
  constraint unit commitment in solar-tidal based microgrids}.
\newblock \emph{\bibinfo{journal}{IET Renewable Power Generation}}
  \bibinfo{year}{2021};.
%Type = Article
\bibitem[{Ikidid et~al.(2021)Ikidid, El~Fazziki and Sadgal}]{ikidid2021multi}
\bibinfo{author}{Ikidid\xfnm[ A.]}, \bibinfo{author}{El~Fazziki\xfnm[ A.]},
  \bibinfo{author}{Sadgal\xfnm[ M.]}.
\newblock \bibinfo{title}{Multi-agent and fuzzy inference-based framework for
  traffic light optimization}.
\newblock \emph{\bibinfo{journal}{International Journal of
  InteractiveMultimedia and Artificial Intelligence}} \bibinfo{year}{2021};.
%Type = Article
\bibitem[{Gomez~Vargas et~al.(2016)Gomez~Vargas, Obreg{\'o}n and
  Alvarez~Pomar}]{gomez2016uncertainty}
\bibinfo{author}{Gomez~Vargas\xfnm[ A.E.]}, \bibinfo{author}{Obreg{\'o}n\xfnm[
  N.]}, \bibinfo{author}{Alvarez~Pomar\xfnm[ C.L.]}.
\newblock \bibinfo{title}{Uncertainty model for quantitative precipitation
  estimation using weather radars}.
\newblock \emph{\bibinfo{journal}{International Journal of Interactive
  Multimedia and Artificial Intelligence}} \bibinfo{year}{2016};.
%Type = Article
\bibitem[{Harish and Roopa(2020)}]{harish2020automated}
\bibinfo{author}{Harish\xfnm[ B.]}, \bibinfo{author}{Roopa\xfnm[ C.]}.
\newblock \bibinfo{title}{Automated ecg analysis for localizing thrombus in
  culprit artery using rule based information fuzzy network}.
\newblock \emph{\bibinfo{journal}{International Journal of Interactive
  Multimedia and Artificial Intelligence}} \bibinfo{year}{2020};.
%Type = Inproceedings
\bibitem[{Elder et~al.(2021)Elder, Arnold, Murthi and
  Navr{\'a}til}]{elder2021learning}
\bibinfo{author}{Elder\xfnm[ B.]}, \bibinfo{author}{Arnold\xfnm[ M.]},
  \bibinfo{author}{Murthi\xfnm[ A.]}, \bibinfo{author}{Navr{\'a}til\xfnm[ J.]}.
\newblock \bibinfo{title}{Learning prediction intervals for model performance}.
\newblock In: \emph{\bibinfo{booktitle}{Proceedings of the AAAI Conference on
  Artificial Intelligence}}; vol.~\bibinfo{volume}{35}.
  \bibinfo{year}{2021}:\unskip \bibinfo{pages}{7305--7313}.
%Type = Inproceedings
\bibitem[{Goel et~al.(2022)Goel, Tung, Eliopoulos, Hu, Thiruvathukal, Davis and
  Lu}]{goel2022directed}
\bibinfo{author}{Goel\xfnm[ A.]}, \bibinfo{author}{Tung\xfnm[ C.]},
  \bibinfo{author}{Eliopoulos\xfnm[ N.]}, \bibinfo{author}{Hu\xfnm[ X.]},
  \bibinfo{author}{Thiruvathukal\xfnm[ G.K.]}, \bibinfo{author}{Davis\xfnm[
  J.C.]}, \bibinfo{author}{Lu\xfnm[ Y.H.]}.
\newblock \bibinfo{title}{Directed acyclic graph-based neural networks for
  tunable low-power computer vision}.
\newblock In: \emph{\bibinfo{booktitle}{Proceedings of the ACM/IEEE
  International Symposium on Low Power Electronics and Design}}.
  \bibinfo{year}{2022}:\unskip \bibinfo{pages}{1--6}.
%Type = Article
\bibitem[{Zilly et~al.(2021)Zilly, Achille, Censi and
  Frazzoli}]{zilly2021plasticity}
\bibinfo{author}{Zilly\xfnm[ J.]}, \bibinfo{author}{Achille\xfnm[ A.]},
  \bibinfo{author}{Censi\xfnm[ A.]}, \bibinfo{author}{Frazzoli\xfnm[ E.]}.
\newblock \bibinfo{title}{On plasticity, invariance, and mutually frozen
  weights in sequential task learning}.
\newblock \emph{\bibinfo{journal}{Advances in Neural Information Processing
  Systems}}
  \bibinfo{year}{2021};\bibinfo{volume}{34}:\bibinfo{pages}{12386--12399}.
%Type = Inproceedings
\bibitem[{Malinin and Gales(2018)}]{malinin2018predictive}
\bibinfo{author}{Malinin\xfnm[ A.]}, \bibinfo{author}{Gales\xfnm[ M.]}.
\newblock \bibinfo{title}{Predictive uncertainty estimation via prior
  networks}.
\newblock In: \emph{\bibinfo{booktitle}{Advances in Neural Information
  Processing Systems}}. \bibinfo{year}{2018}:\unskip
  \bibinfo{pages}{7047--7058}.
%Type = Article
\bibitem[{Ma et~al.(2022)Ma, Wu, Li, Guo, Jiang, Zhu and Wu}]{ma2022hw}
\bibinfo{author}{Ma\xfnm[ P.]}, \bibinfo{author}{Wu\xfnm[ Y.]},
  \bibinfo{author}{Li\xfnm[ Y.]}, \bibinfo{author}{Guo\xfnm[ L.]},
  \bibinfo{author}{Jiang\xfnm[ H.]}, \bibinfo{author}{Zhu\xfnm[ X.]},
  \bibinfo{author}{Wu\xfnm[ X.]}.
\newblock \bibinfo{title}{Hw-forest: Deep forest with hashing screening and
  window screening}.
\newblock \emph{\bibinfo{journal}{ACM Transactions on Knowledge Discovery from
  Data (TKDD)}} \bibinfo{year}{2022};.
%Type = Article
\bibitem[{ITO et~al.(2022)ITO, HE and OKI}]{ito2022backup}
\bibinfo{author}{ITO\xfnm[ M.]}, \bibinfo{author}{HE\xfnm[ F.]},
  \bibinfo{author}{OKI\xfnm[ E.]}.
\newblock \bibinfo{title}{Backup resource allocation of virtual machines for
  probabilistic protection under capacity uncertainty}.
\newblock \emph{\bibinfo{journal}{IEICE Transactions on Communications}}
  \bibinfo{year}{2022};:\bibinfo{pages}{2021EBP3144}.
%Type = Article
\bibitem[{Siljak et~al.(2021)Siljak, Macaluso and
  Marchetti}]{siljak2021artificial}
\bibinfo{author}{Siljak\xfnm[ H.]}, \bibinfo{author}{Macaluso\xfnm[ I.]},
  \bibinfo{author}{Marchetti\xfnm[ N.]}.
\newblock \bibinfo{title}{Artificial intelligence for dynamical systems in
  wireless communications: Modeling for the future}.
\newblock \emph{\bibinfo{journal}{IEEE Systems, Man, and Cybernetics Magazine}}
  \bibinfo{year}{2021};\bibinfo{volume}{7}(\bibinfo{number}{4}):\bibinfo{pages}{13--23}.
%Type = Inproceedings
\bibitem[{Kabir et~al.(2020{\natexlab{a}})Kabir, Khosravi, Nahavandi and
  Nahavandi}]{kabir2020uncertainty}
\bibinfo{author}{Kabir\xfnm[ H.D.]}, \bibinfo{author}{Khosravi\xfnm[ A.]},
  \bibinfo{author}{Nahavandi\xfnm[ D.]}, \bibinfo{author}{Nahavandi\xfnm[ S.]}.
\newblock \bibinfo{title}{Uncertainty quantification neural network from
  similarity and sensitivity}.
\newblock In: \emph{\bibinfo{booktitle}{2020 International Joint Conference on
  Neural Networks (IJCNN)}}. \bibinfo{organization}{IEEE};
  \bibinfo{year}{2020}{\natexlab{a}}:\unskip \bibinfo{pages}{1--8}.
%Type = Article
\bibitem[{Abdar et~al.(2021)Abdar, Pourpanah, Hussain, Rezazadegan, Liu,
  Ghavamzadeh, Fieguth, Cao, Khosravi, Acharya et~al.}]{abdar2021review}
\bibinfo{author}{Abdar\xfnm[ M.]}, \bibinfo{author}{Pourpanah\xfnm[ F.]},
  \bibinfo{author}{Hussain\xfnm[ S.]}, \bibinfo{author}{Rezazadegan\xfnm[ D.]},
  \bibinfo{author}{Liu\xfnm[ L.]}, \bibinfo{author}{Ghavamzadeh\xfnm[ M.]},
  \bibinfo{author}{Fieguth\xfnm[ P.]}, \bibinfo{author}{Cao\xfnm[ X.]},
  \bibinfo{author}{Khosravi\xfnm[ A.]}, \bibinfo{author}{Acharya\xfnm[ U.R.]},
  et~al.
\newblock \bibinfo{title}{A review of uncertainty quantification in deep
  learning: Techniques, applications and challenges}.
\newblock \emph{\bibinfo{journal}{Information Fusion}}
  \bibinfo{year}{2021};\bibinfo{volume}{76}:\bibinfo{pages}{243--297}.
%Type = Inproceedings
\bibitem[{Gal and Ghahramani(2016)}]{gal2016dropout}
\bibinfo{author}{Gal\xfnm[ Y.]}, \bibinfo{author}{Ghahramani\xfnm[ Z.]}.
\newblock \bibinfo{title}{Dropout as a bayesian approximation: Representing
  model uncertainty in deep learning}.
\newblock In: \emph{\bibinfo{booktitle}{international conference on machine
  learning}}. \bibinfo{organization}{PMLR}; \bibinfo{year}{2016}:\unskip
  \bibinfo{pages}{1050--1059}.
%Type = Article
\bibitem[{Kabir et~al.(2018)Kabir, Khosravi, Hosen and
  Nahavandi}]{kabir2018neural}
\bibinfo{author}{Kabir\xfnm[ H.D.]}, \bibinfo{author}{Khosravi\xfnm[ A.]},
  \bibinfo{author}{Hosen\xfnm[ M.A.]}, \bibinfo{author}{Nahavandi\xfnm[ S.]}.
\newblock \bibinfo{title}{Neural network-based uncertainty quantification: A
  survey of methodologies and applications}.
\newblock \emph{\bibinfo{journal}{IEEE Access}} \bibinfo{year}{2018};.
%Type = Article
\bibitem[{Ajlan et~al.(2022)Ajlan, Rahim, Ismael et~al.}]{ajlan2022text}
\bibinfo{author}{Ajlan\xfnm[ I.K.]}, \bibinfo{author}{Rahim\xfnm[ H.B.A.]},
  \bibinfo{author}{Ismael\xfnm[ A.J.]}, et~al.
\newblock \bibinfo{title}{Text recognition from images}.
\newblock \emph{\bibinfo{journal}{Texas Journal of Engineering and Technology}}
  \bibinfo{year}{2022};\bibinfo{volume}{10}:\bibinfo{pages}{10--18}.
%Type = Article
\bibitem[{Kamal et~al.(2021)Kamal, Northcote, Chowdhury, Dey, Crespo and
  Herrera-Viedma}]{kamal2021alzheimer}
\bibinfo{author}{Kamal\xfnm[ M.S.]}, \bibinfo{author}{Northcote\xfnm[ A.]},
  \bibinfo{author}{Chowdhury\xfnm[ L.]}, \bibinfo{author}{Dey\xfnm[ N.]},
  \bibinfo{author}{Crespo\xfnm[ R.G.]}, \bibinfo{author}{Herrera-Viedma\xfnm[
  E.]}.
\newblock \bibinfo{title}{Alzheimer’s patient analysis using image and gene
  expression data and explainable-ai to present associated genes}.
\newblock \emph{\bibinfo{journal}{IEEE Transactions on Instrumentation and
  Measurement}}
  \bibinfo{year}{2021};\bibinfo{volume}{70}:\bibinfo{pages}{1--7}.
%Type = Article
\bibitem[{Campbell and Beronov(2019)}]{campbell2019sparse}
\bibinfo{author}{Campbell\xfnm[ T.]}, \bibinfo{author}{Beronov\xfnm[ B.]}.
\newblock \bibinfo{title}{Sparse variational inference: Bayesian coresets from
  scratch}.
\newblock \emph{\bibinfo{journal}{Advances in Neural Information Processing
  Systems}} \bibinfo{year}{2019};\bibinfo{volume}{32}.
%Type = Article
\bibitem[{Zhou et~al.(2021)Zhou, Liu, Pourpanah, Zeng and
  Wang}]{zhou2021survey}
\bibinfo{author}{Zhou\xfnm[ X.]}, \bibinfo{author}{Liu\xfnm[ H.]},
  \bibinfo{author}{Pourpanah\xfnm[ F.]}, \bibinfo{author}{Zeng\xfnm[ T.]},
  \bibinfo{author}{Wang\xfnm[ X.]}.
\newblock \bibinfo{title}{A survey on epistemic (model) uncertainty in
  supervised learning: Recent advances and applications}.
\newblock \emph{\bibinfo{journal}{Neurocomputing}} \bibinfo{year}{2021};.
%Type = Article
\bibitem[{Olivier et~al.(2021)Olivier, Shields and
  Graham-Brady}]{olivier2021bayesian}
\bibinfo{author}{Olivier\xfnm[ A.]}, \bibinfo{author}{Shields\xfnm[ M.D.]},
  \bibinfo{author}{Graham-Brady\xfnm[ L.]}.
\newblock \bibinfo{title}{Bayesian neural networks for uncertainty
  quantification in data-driven materials modeling}.
\newblock \emph{\bibinfo{journal}{Computer Methods in Applied Mechanics and
  Engineering}}
  \bibinfo{year}{2021};\bibinfo{volume}{386}:\bibinfo{pages}{114079}.
%Type = Inproceedings
\bibitem[{Serpell et~al.(2019)Serpell, Araya, Valle and
  Allende}]{serpell2019probabilistic}
\bibinfo{author}{Serpell\xfnm[ C.]}, \bibinfo{author}{Araya\xfnm[ I.]},
  \bibinfo{author}{Valle\xfnm[ C.]}, \bibinfo{author}{Allende\xfnm[ H.]}.
\newblock \bibinfo{title}{Probabilistic forecasting using monte carlo dropout
  neural networks}.
\newblock In: \emph{\bibinfo{booktitle}{Iberoamerican Congress on Pattern
  Recognition}}. \bibinfo{organization}{Springer}; \bibinfo{year}{2019}:\unskip
  \bibinfo{pages}{387--397}.
%Type = Article
\bibitem[{Posch and Pilz(2020)}]{posch2020correlated}
\bibinfo{author}{Posch\xfnm[ K.]}, \bibinfo{author}{Pilz\xfnm[ J.]}.
\newblock \bibinfo{title}{Correlated parameters to accurately measure
  uncertainty in deep neural networks}.
\newblock \emph{\bibinfo{journal}{IEEE Transactions on Neural Networks and
  Learning Systems}}
  \bibinfo{year}{2020};\bibinfo{volume}{32}(\bibinfo{number}{3}):\bibinfo{pages}{1037--1051}.
%Type = Article
\bibitem[{Jospin et~al.(2022)Jospin, Laga, Boussaid, Buntine and
  Bennamoun}]{jospin2022hands}
\bibinfo{author}{Jospin\xfnm[ L.V.]}, \bibinfo{author}{Laga\xfnm[ H.]},
  \bibinfo{author}{Boussaid\xfnm[ F.]}, \bibinfo{author}{Buntine\xfnm[ W.]},
  \bibinfo{author}{Bennamoun\xfnm[ M.]}.
\newblock \bibinfo{title}{Hands-on bayesian neural networks—a tutorial for
  deep learning users}.
\newblock \emph{\bibinfo{journal}{IEEE Computational Intelligence Magazine}}
  \bibinfo{year}{2022};\bibinfo{volume}{17}(\bibinfo{number}{2}):\bibinfo{pages}{29--48}.
%Type = Article
\bibitem[{Upadhyay et~al.(2021)Upadhyay, Chen and
  Akata}]{upadhyay2021robustness}
\bibinfo{author}{Upadhyay\xfnm[ U.]}, \bibinfo{author}{Chen\xfnm[ Y.]},
  \bibinfo{author}{Akata\xfnm[ Z.]}.
\newblock \bibinfo{title}{Robustness via uncertainty-aware cycle consistency}.
\newblock \emph{\bibinfo{journal}{Advances in Neural Information Processing
  Systems}}
  \bibinfo{year}{2021};\bibinfo{volume}{34}:\bibinfo{pages}{28261--28273}.
%Type = Article
\bibitem[{Khosravi et~al.(2010)Khosravi, Nahavandi, Creighton and
  Atiya}]{khosravi2010lower}
\bibinfo{author}{Khosravi\xfnm[ A.]}, \bibinfo{author}{Nahavandi\xfnm[ S.]},
  \bibinfo{author}{Creighton\xfnm[ D.]}, \bibinfo{author}{Atiya\xfnm[ A.F.]}.
\newblock \bibinfo{title}{Lower upper bound estimation method for construction
  of neural network-based prediction intervals}.
\newblock \emph{\bibinfo{journal}{IEEE transactions on neural networks}}
  \bibinfo{year}{2010};\bibinfo{volume}{22}(\bibinfo{number}{3}):\bibinfo{pages}{337--346}.
%Type = Article
\bibitem[{Khosravi et~al.(2011)Khosravi, Nahavandi, Creighton and
  Atiya}]{khosravi2011comprehensive}
\bibinfo{author}{Khosravi\xfnm[ A.]}, \bibinfo{author}{Nahavandi\xfnm[ S.]},
  \bibinfo{author}{Creighton\xfnm[ D.]}, \bibinfo{author}{Atiya\xfnm[ A.F.]}.
\newblock \bibinfo{title}{Comprehensive review of neural network-based
  prediction intervals and new advances}.
\newblock \emph{\bibinfo{journal}{IEEE Transactions on neural networks}}
  \bibinfo{year}{2011};\bibinfo{volume}{22}(\bibinfo{number}{9}):\bibinfo{pages}{1341--1356}.
%Type = Article
\bibitem[{Kabir et~al.(2021)Kabir, Khosravi, Kavousi-Fard, Nahavandi and
  Srinivasan}]{kabir2021optimal}
\bibinfo{author}{Kabir\xfnm[ H.D.]}, \bibinfo{author}{Khosravi\xfnm[ A.]},
  \bibinfo{author}{Kavousi-Fard\xfnm[ A.]}, \bibinfo{author}{Nahavandi\xfnm[
  S.]}, \bibinfo{author}{Srinivasan\xfnm[ D.]}.
\newblock \bibinfo{title}{Optimal uncertainty-guided neural network training}.
\newblock \emph{\bibinfo{journal}{Applied Soft Computing}}
  \bibinfo{year}{2021};\bibinfo{volume}{99}:\bibinfo{pages}{106878}.
%Type = Inproceedings
\bibitem[{Mar{\'\i}n et~al.(2016)Mar{\'\i}n, Valencia and
  S{\'a}ez}]{marin2016prediction}
\bibinfo{author}{Mar{\'\i}n\xfnm[ L.G.]}, \bibinfo{author}{Valencia\xfnm[ F.]},
  \bibinfo{author}{S{\'a}ez\xfnm[ D.]}.
\newblock \bibinfo{title}{Prediction interval based on type-2 fuzzy systems for
  wind power generation and loads in microgrid control design}.
\newblock In: \emph{\bibinfo{booktitle}{Fuzzy Systems (FUZZ-IEEE), 2016 IEEE
  International Conference on}}. \bibinfo{organization}{IEEE};
  \bibinfo{year}{2016}:\unskip \bibinfo{pages}{328--335}.
%Type = Article
\bibitem[{Zhang et~al.(2015)Zhang, Wu, Wong, Xu, Dong and
  Iu}]{zhang2015advanced}
\bibinfo{author}{Zhang\xfnm[ G.]}, \bibinfo{author}{Wu\xfnm[ Y.]},
  \bibinfo{author}{Wong\xfnm[ K.P.]}, \bibinfo{author}{Xu\xfnm[ Z.]},
  \bibinfo{author}{Dong\xfnm[ Z.Y.]}, \bibinfo{author}{Iu\xfnm[ H.H.C.]}.
\newblock \bibinfo{title}{An advanced approach for construction of optimal wind
  power prediction intervals}.
\newblock \emph{\bibinfo{journal}{IEEE Transactions on Power Systems}}
  \bibinfo{year}{2015};\bibinfo{volume}{30}(\bibinfo{number}{5}):\bibinfo{pages}{2706--2715}.
%Type = Article
\bibitem[{Burnstein et~al.(1961)Burnstein, Stotland and
  Zander}]{burnstein1961similarity}
\bibinfo{author}{Burnstein\xfnm[ E.]}, \bibinfo{author}{Stotland\xfnm[ E.]},
  \bibinfo{author}{Zander\xfnm[ A.]}.
\newblock \bibinfo{title}{Similarity to a model and self-evaluation.}
\newblock \emph{\bibinfo{journal}{The Journal of Abnormal and Social
  Psychology}}
  \bibinfo{year}{1961};\bibinfo{volume}{62}(\bibinfo{number}{2}):\bibinfo{pages}{257}.
%Type = Article
\bibitem[{Hiza and Duncan(1970)}]{hiza1970correlation}
\bibinfo{author}{Hiza\xfnm[ M.]}, \bibinfo{author}{Duncan\xfnm[ A.]}.
\newblock \bibinfo{title}{A correlation for the prediction of interaction
  energy parameters for mixtures of small molecules}.
\newblock \emph{\bibinfo{journal}{AIChE Journal}}
  \bibinfo{year}{1970};\bibinfo{volume}{16}(\bibinfo{number}{5}):\bibinfo{pages}{733--738}.
%Type = Inproceedings
\bibitem[{Yao et~al.(2019)Yao, Tang, Wei, Zheng and Li}]{yao2019revisiting}
\bibinfo{author}{Yao\xfnm[ H.]}, \bibinfo{author}{Tang\xfnm[ X.]},
  \bibinfo{author}{Wei\xfnm[ H.]}, \bibinfo{author}{Zheng\xfnm[ G.]},
  \bibinfo{author}{Li\xfnm[ Z.]}.
\newblock \bibinfo{title}{Revisiting spatial-temporal similarity: A deep
  learning framework for traffic prediction}.
\newblock In: \emph{\bibinfo{booktitle}{Proceedings of the AAAI conference on
  artificial intelligence}}; vol.~\bibinfo{volume}{33}.
  \bibinfo{year}{2019}:\unskip \bibinfo{pages}{5668--5675}.
%Type = Article
\bibitem[{Khan and Aziz(2011)}]{khan2011natural}
\bibinfo{author}{Khan\xfnm[ W.]}, \bibinfo{author}{Aziz\xfnm[ A.]}.
\newblock \bibinfo{title}{Natural convection flow of a nanofluid over a
  vertical plate with uniform surface heat flux}.
\newblock \emph{\bibinfo{journal}{International Journal of Thermal Sciences}}
  \bibinfo{year}{2011};\bibinfo{volume}{50}(\bibinfo{number}{7}):\bibinfo{pages}{1207--1214}.
%Type = Article
\bibitem[{VanDecar and Crosson(1990)}]{vandecar1990determination}
\bibinfo{author}{VanDecar\xfnm[ J.]}, \bibinfo{author}{Crosson\xfnm[ R.]}.
\newblock \bibinfo{title}{Determination of teleseismic relative phase arrival
  times using multi-channel cross-correlation and least squares}.
\newblock \emph{\bibinfo{journal}{Bulletin of the Seismological Society of
  America}}
  \bibinfo{year}{1990};\bibinfo{volume}{80}(\bibinfo{number}{1}):\bibinfo{pages}{150--169}.
%Type = Article
\bibitem[{Kabir et~al.(2020{\natexlab{b}})Kabir, Khosravi, Nahavandi and
  Srinivasan}]{kabir2020neural}
\bibinfo{author}{Kabir\xfnm[ H.D.]}, \bibinfo{author}{Khosravi\xfnm[ A.]},
  \bibinfo{author}{Nahavandi\xfnm[ S.]}, \bibinfo{author}{Srinivasan\xfnm[
  D.]}.
\newblock \bibinfo{title}{Neural network training for uncertainty
  quantification over time-range}.
\newblock \emph{\bibinfo{journal}{IEEE Transactions on Emerging Topics in
  Computational Intelligence}}
  \bibinfo{year}{2020}{\natexlab{b}};\bibinfo{volume}{5}(\bibinfo{number}{5}):\bibinfo{pages}{768--779}.
%Type = Article
\bibitem[{Kabir et~al.(2023)Kabir, Abdar, Khosravi, Nahavandi, Mondal, Khanam,
  Mohamed, Srinivasan, Nahavandi and Suganthan}]{kabir2023synthetic}
\bibinfo{author}{Kabir\xfnm[ H.D.]}, \bibinfo{author}{Abdar\xfnm[ M.]},
  \bibinfo{author}{Khosravi\xfnm[ A.]}, \bibinfo{author}{Nahavandi\xfnm[ D.]},
  \bibinfo{author}{Mondal\xfnm[ S.K.]}, \bibinfo{author}{Khanam\xfnm[ S.]},
  \bibinfo{author}{Mohamed\xfnm[ S.]}, \bibinfo{author}{Srinivasan\xfnm[ D.]},
  \bibinfo{author}{Nahavandi\xfnm[ S.]}, \bibinfo{author}{Suganthan\xfnm[
  P.N.]}.
\newblock \bibinfo{title}{Synthetic datasets for numeric uncertainty
  quantification: Proposing datasets for future researchers}.
\newblock \emph{\bibinfo{journal}{IEEE Systems, Man, and Cybernetics Magazine}}
  \bibinfo{year}{2023};\bibinfo{volume}{9}(\bibinfo{number}{2}):\bibinfo{pages}{39--48}.
%Type = Misc
\bibitem[{Jarshi(2018)}]{LE_DATA}
\bibinfo{author}{Jarshi\xfnm[ K.]}.
\newblock \bibinfo{title}{Life expectancy (who): Statistical analysis on
  factors influencing life expectancy}.
\newblock \bibinfo{year}{2018}.
\newblock \urlprefix\url{www.kaggle.com
  /datasets/kumarajarshi/life-expectancy-who}.
%Type = Inproceedings
\bibitem[{Blundell et~al.(2015)Blundell, Cornebise, Kavukcuoglu and
  Wierstra}]{blundell2015weight}
\bibinfo{author}{Blundell\xfnm[ C.]}, \bibinfo{author}{Cornebise\xfnm[ J.]},
  \bibinfo{author}{Kavukcuoglu\xfnm[ K.]}, \bibinfo{author}{Wierstra\xfnm[
  D.]}.
\newblock \bibinfo{title}{Weight uncertainty in neural network}.
\newblock In: \emph{\bibinfo{booktitle}{International conference on machine
  learning}}. \bibinfo{organization}{PMLR}; \bibinfo{year}{2015}:\unskip
  \bibinfo{pages}{1613--1622}.
%Type = Misc
\bibitem[{Uber~Technologies(2018)}]{Pyro}
\bibinfo{author}{Uber~Technologies\xfnm[ I.]}.
\newblock \bibinfo{title}{Bayesian regression tutorial}.
\newblock \bibinfo{year}{2018}.
\newblock \urlprefix\url{pyro.ai/examples/bayesian\_regression.html}.
%Type = Article
\bibitem[{Lee et~al.(2022)Lee, Kim and Lee}]{lee2022graddiv}
\bibinfo{author}{Lee\xfnm[ S.]}, \bibinfo{author}{Kim\xfnm[ H.]},
  \bibinfo{author}{Lee\xfnm[ J.]}.
\newblock \bibinfo{title}{Graddiv: Adversarial robustness of randomized neural
  networks via gradient diversity regularization}.
\newblock \emph{\bibinfo{journal}{IEEE Transactions on Pattern Analysis and
  Machine Intelligence}} \bibinfo{year}{2022};.

\end{thebibliography}

\end{document}